%% file: main_arxiv.tex
\PassOptionsToPackage{table}{xcolor}
\documentclass[11pt]{article}

\usepackage[preprint]{acl}

\usepackage{times}
\usepackage{latexsym}

\usepackage[T1]{fontenc}

\usepackage[utf8]{inputenc}

\usepackage{microtype}

\usepackage{inconsolata}

\usepackage{graphicx}
\usepackage{xcolor}
\usepackage{booktabs}
\usepackage{multirow}
\usepackage{colortbl}
\usepackage{amsmath}
\usepackage{subcaption}
\usepackage{placeins}

\title{FADE: Mitigating Hallucinations by Reducing Language-Prior Dominance in Large Vision-Language Models}

\author{
  \textbf{Yichen Guo\textsuperscript{1,2,*}} \quad
  \textbf{Kai Tang\textsuperscript{1,2,*}} \quad
  \textbf{Jinhao You\textsuperscript{4,*}} \\
  \textbf{Fenglai Lin\textsuperscript{1}} \quad
  \textbf{Yiding Sun\textsuperscript{2}} \quad
  \textbf{Dongxu Zhang\textsuperscript{3}} \\
  \textbf{Wenya Wang\textsuperscript{1}} \quad
  \textbf{Lin William Cong\textsuperscript{1}} \quad
  \textbf{Shanghang Zhang\textsuperscript{2,\dag}} \\
  \textsuperscript{1}Nanyang Technological University \\
  \textsuperscript{2}State Key Laboratory of Multimedia Information Processing, School of Computer Science, Peking University \\
  \textsuperscript{3}Tsinghua University \\
  \textsuperscript{4}University of Pennsylvania \\
  \small{\textsuperscript{*}Equal contribution. \quad \textsuperscript{\dag}Corresponding author: \href{mailto:shanghang@pku.edu.cn}{shanghang@pku.edu.cn}}
}

\begin{document}
\setlength{\abovedisplayskip}{3pt plus 1pt minus 1.5pt}
\setlength{\belowdisplayskip}{3pt plus 1pt minus 1.5pt}
\setlength{\abovedisplayshortskip}{0pt plus 1pt minus 1.5pt}
\setlength{\belowdisplayshortskip}{2pt plus 1pt minus 1.5pt}
\maketitle
\begin{abstract}

Despite the impressive capabilities of Large Vision-Language Models (LVLMs), they remain susceptible to hallucination, generating content inconsistent with the input image. Recent studies attribute this to the dominance of language priors over visual inputs and employ contrastive decoding methods to mitigate this dominance, but the mechanistic origin remains unexplored. We investigate the information flow through each transformer layer and find that attention modules consistently aggregate visual evidence, while FFN modules at critical layers act as the source of language priors. These priors can override visual evidence, causing correct predictions in intermediate layers to drift toward incorrect outputs. Based on this insight, we propose \textbf{FADE} (\textbf{F}FN \textbf{A}ttenuation for \textbf{DE}coding), a training-free method that attenuates FFN outputs to reduce language-prior dominance. Evaluations on POPE, CHAIR, and MME benchmarks across LLaVA-1.5, mPLUG-Owl2, and InstructBLIP show that FADE effectively mitigates hallucinations while preserving inference efficiency. Code is available at \url{https://github.com/EasonAI-5589/LLaVA-Hallucination}.

\end{abstract}

\input{sections/1_introduction_updated_ref}
\input{sections/2_related_work_updated_ref}
\input{sections/3_method_updated_ref}
\input{sections/4_experiments_updated_ref}
\input{sections/5_conclusion}

\section*{Limitations}
Our main experiments focus on 7B-scale models; while we report 13B results in Appendix~\ref{sec:appendix_13b} showing consistent generalization, extending to 30B+ scales remains future work. Our evaluation emphasizes hallucination-specific benchmarks (POPE, CHAIR, MME), so performance on broader VQA or reasoning tasks is untested. The critical layer is fixed per model architecture; exploring adaptive layer selection is a promising direction.

\section*{Acknowledgments}
This work was supported by the National Natural Science Foundation of China (62476011), and by the Beijing Natural Science Foundation (L252060).

\bibliography{references}

\clearpage
\input{sections/appendix_updated_ref}

\end{document}

%% file: sections/1_introduction_updated_ref.tex
\section{Introduction}
\label{sec:introduction}

\begin{figure}[t]
\centering
\includegraphics[width=\linewidth]{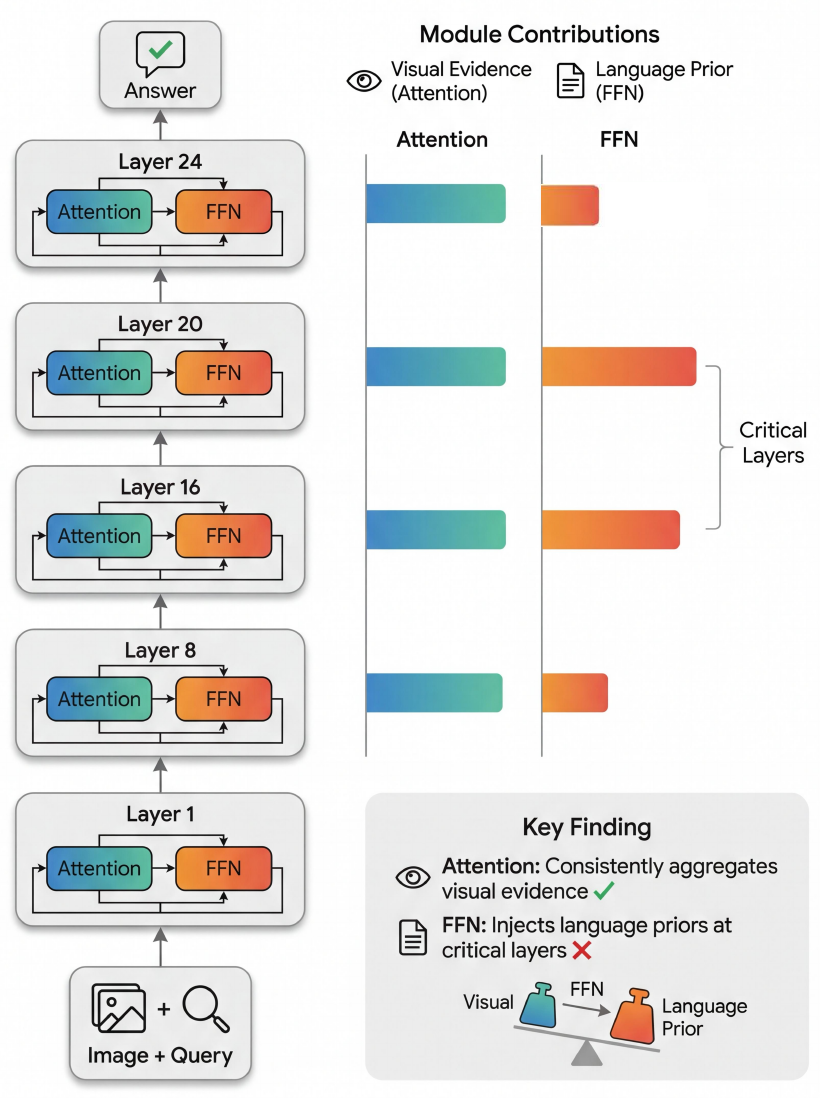}
\caption{Analyzing information flow through transformer layers. Attention consistently aggregates visual evidence, while FFN at critical layers (16--22) introduces language priors that can override visual evidence.}
\label{fig:main}
\vspace{-5mm}
\end{figure}

Large Vision-Language Models (LVLMs) have achieved remarkable progress in recent years, bridging the gap between vision and language through effective multimodal alignment~\cite{radford2021learning,li2022blip,li2023blip,liu2023visual,chen2024internvl,bai2023qwenvl,wang2024qwen2}. These models have achieved significant success across diverse applications including visual question answering (VQA), image captioning, and multimodal geometric reasoning~\cite{zhang2026pointcot}. However, a persistent challenge remains: LVLMs often generate text that is not consistent with the visual content of the input image, known as hallucination~\cite{li2023evaluating,rohrbach2018object,liu2024surveyhallucinationlargevisionlanguage,bai2025hallucinationmultimodallargelanguage}. This phenomenon can cause serious risks in critical applications, including medical diagnosis, autonomous driving~\cite{cui2024survey}, and embodied agents~\cite{driess2023palm}, where precision and reliability are essential.

\begin{figure*}[t]
    \centering
    \includegraphics[width=0.9\textwidth]{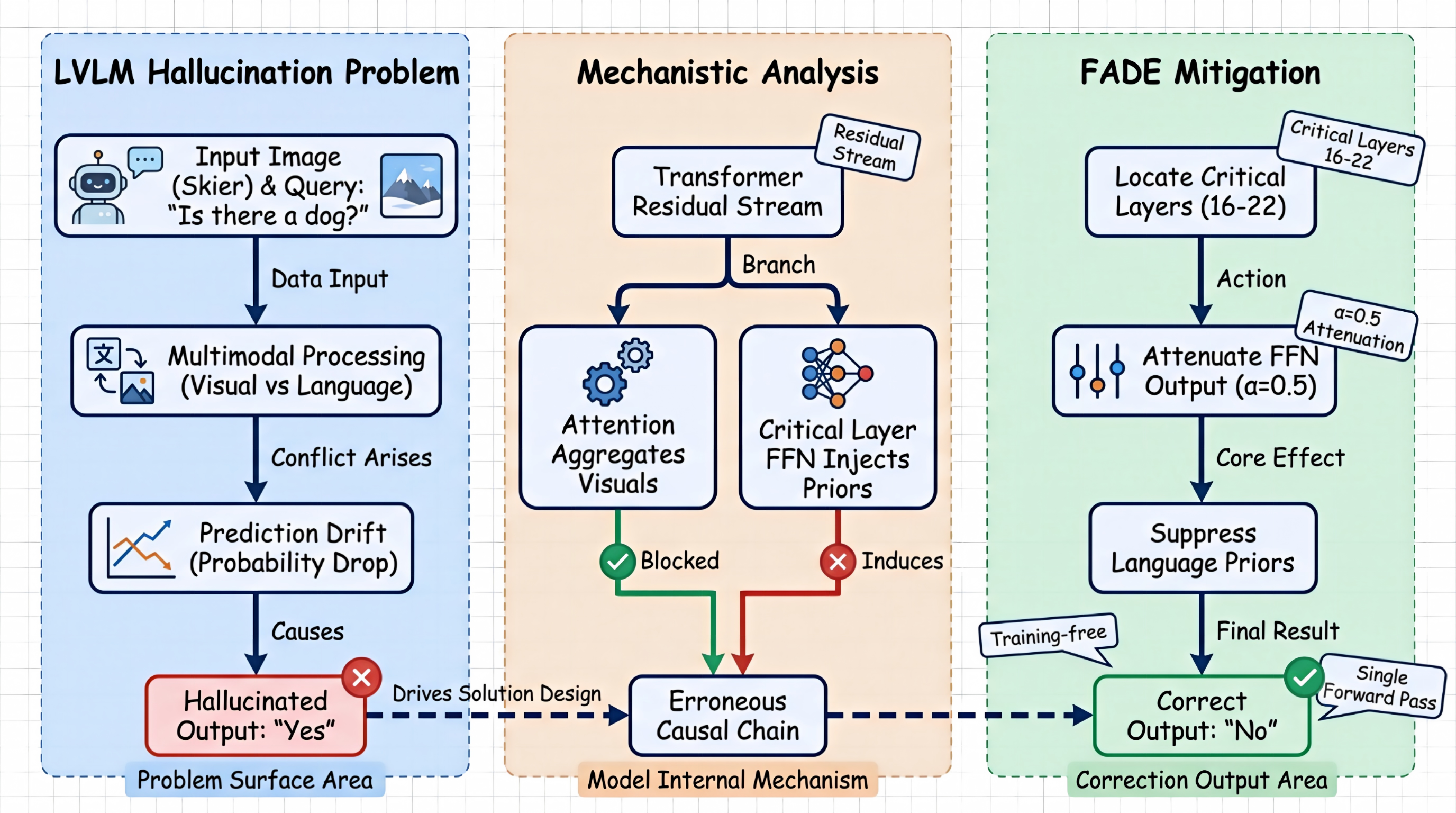}
    \caption{Overview of our approach. \textbf{Left:} LVLMs suffer from hallucinations where language priors override visual evidence, causing prediction drift from correct to incorrect outputs. \textbf{Middle:} Our mechanistic analysis reveals that attention modules aggregate visual evidence toward correct answers, while FFN modules at critical layers introduce language priors that can override visual evidence. \textbf{Right:} FADE attenuates FFN outputs at critical layers to suppress language priors while preserving visual evidence, enabling training-free hallucination mitigation.}
    \label{fig:overview}
    \vspace{-4mm}
\end{figure*}

Recent research on mitigating hallucinations can be divided into two categories. Training-based approaches employ instruction tuning~\cite{liu2024improved}, RLHF~\cite{sun2024aligning} or DPO~\cite{zhao2023beyond} to reduce hallucinations at the source, but they require expensive data collection and retraining. Training-free methods intervene during inference without modifying model parameters. Attention modification approaches~\cite{liu2024paying,huang2024opera} amplify visual token weights to enhance visual grounding. Layer-wise intervention methods~\cite{chuang2024dola,wang2025damo} exploit cross-layer differences to improve output quality. Contrastive decoding methods~\cite{leng2024mitigating,manevich2024mitigating} attribute hallucination to the dominance of language priors over visual inputs and attempt to suppress this dominance by contrasting output distributions. However, these methods operate at the output level without understanding where language priors originate within the model. Understanding this origin is crucial for developing more targeted and efficient solutions.

In this work, we investigate the mechanistic origin of language-prior dominance. We decompose transformer computations into attention and FFN contributions using the residual stream perspective~\cite{elhage2021mathematical}, and measure their effects on predictions through logit lens projections~\cite{geva2022transformer,belrose2023eliciting}. As illustrated in Figure~\ref{fig:main}, our analysis reveals two key findings: (1) \emph{Attention Aggregates Visual Evidence.} Attention mechanisms consistently aggregate visual features to generate correct predictions. (2) \emph{FFN Introduces Language Priors.} FFN modules at critical layers act as the source of language priors that can override visual evidence, causing hallucinations.

Based on this insight, we propose \textbf{FADE} (\textbf{F}FN \textbf{A}ttenuation for \textbf{DE}coding), a training-free method that attenuates FFN outputs at critical layers to reduce language-prior dominance (Figure~\ref{fig:overview}). By weakening FFN contributions, FADE preserves the visual evidence while suppressing the language priors that cause hallucination. Unlike contrastive decoding, FADE operates in a single forward pass with minimal~overhead.

Our contributions can be summarized as follows:
\begin{itemize}
    \item We conduct a mechanistic analysis revealing the origin of language prior dominance in LVLMs: attention aggregates visual evidence, while FFN at critical layers acts as the source of language priors that can override it.
    \item We propose FADE, a training-free method that attenuates FFN outputs at critical layers to reduce language-prior dominance while preserving visual evidence.
    \item Extensive experiments across diverse architectures (LLaVA-1.5-7B/13B, mPLUG-Owl2, InstructBLIP, InternVL3-8B, Qwen2.5/3-VL) and six benchmarks (POPE, CHAIR, MME, MMHal-Bench, HalBench, MMBench) demonstrate that FADE effectively mitigates hallucinations while maintaining inference efficiency and general capabilities.
\end{itemize}

%% file: sections/2_related_work_updated_ref.tex
\section{Related Work}
\label{sec:related_work}

\subsection{Large Vision-Language Models}
\label{subsec:lvlms}
Large Vision-Language Models (LVLMs) have evolved from early BERT-based decoders~\cite{chen2020uniter, li2020oscar, zhang2021vinvl, li2021align, wang2021simvlm, li2022blip} designed to integrate visual and textual information into a paradigm driven by large language models (LLMs)~\cite{touvron2023llama, touvron2023llama2, jiang2023mistral7b, grattafiori2024llama, wang2024qwen2}. The emergence of LLMs has sustainably enhanced the capabilities and performance of LVLMs. In this process, supported by end-to-end training techniques~\cite{alayrac2022flamingo, dai2023instructblip}, LVLMs have achieved unified decoding of visual and textual tokens, indicating that both their expressiveness and adaptability have significantly improved. Recent works, such as LLaVA~\cite{liu2023visual, liu2024improved, liu2024llavanext} and InstructBLIP~\cite{dai2023instructblip}, have further refined these models through visual instruction tuning, enhancing their performance in various vision-language tasks. More recently, models such as the Qwen-VL series~\cite{bai2023qwenvl, wang2024qwen2} and InternVL series~\cite{chen2024internvl, chen2024far} have further scaled up through improved alignment strategies and large-scale joint training.

\subsection{Hallucination Mitigation in LVLMs}
\label{subsec:hallucination}

Hallucination in LVLMs refers to generating content that is linguistically plausible but inconsistent with visual input~\cite{rohrbach2018object, li2023evaluating}.
Training-based approaches mitigate it via additional fine-tuning---robust instruction tuning~\cite{liu2024mitigating}, post-hoc revision~\cite{zhou2024analyzing}, RLHF~\cite{yu2024rlhf}, or DPO~\cite{zhao2023beyond, wang2024vigc}---but incur substantial training costs.

Training-free methods instead operate during inference. \emph{Attention-based methods} re-weight attention to strengthen visual grounding (PAI~\cite{liu2024paying}, OPERA~\cite{huang2024opera}, AGLA/All-Path~\cite{an2025mitigating, qian2026interveneallpathsunifiedmitigationlvlm}). \emph{Contrastive decoding} suppresses hallucinated content by contrasting distributions from original and perturbed inputs~\cite{leng2024mitigating,manevich2024mitigating}, instructions~\cite{wang2024mitigating}, or self-generated descriptions~\cite{kim2024code}. \emph{Layer-wise intervention} exploits the transformer hierarchy: DAMO~\cite{wang2025damo} accumulates activation momentum, while others contrast logits across layers~\cite{chuang2024dola} or enforce inter-layer consistency~\cite{huo2025self, li2025map, tang2025mitigating}. \emph{Representation engineering} manipulates hidden states via pre-computed steering vectors (VISTA~\cite{li2025hidden}, VTI~\cite{liu2025reducing}, FlexAC~\cite{lyu2026flexac}).

Concurrent work on layer-wise transformer dynamics includes \citet{neo2025towards}, which analyzes visual-token processing via attention knockouts but proposes no hallucination mitigation, and ReDeEP~\cite{sun2025redeep}, which targets retrieval-augmented generation and requires \emph{dual} intervention because attention fails to retain external context there. In contrast, our contrastive analysis on vision-language hallucination shows that attention remains reliable across correct and hallucinated samples, while FFN at critical layers is the divergence point---motivating \textbf{FADE} (\textbf{F}FN \textbf{A}ttenuation for \textbf{DE}coding), a training-free single-component intervention that attenuates FFN outputs at those layers to reduce language-prior dominance.

%% file: sections/3_method_updated_ref.tex
\section{Method}
\label{sec:method}

\subsection{Preliminaries}
\label{subsec:preliminaries}

A transformer-based LVLM processes inputs through $L$ decoder layers. Each layer $l$ applies attention and FFN with residual connections:
\begin{align}
    \tilde{\mathbf{h}}^{(l)} &= \mathbf{h}^{(l)} + \mathrm{Attn}^{(l)}(\mathbf{h}^{(l)}) \\
    \mathbf{h}^{(l+1)} &= \tilde{\mathbf{h}}^{(l)} + \mathrm{FFN}^{(l)}(\tilde{\mathbf{h}}^{(l)})
\end{align}
From the residual stream perspective \cite{elhage2021mathematical}, attention aggregates information across positions while FFN performs per-position transformations. Prior work shows FFN layers function as key-value memories storing factual knowledge \cite{geva2021transformer, meng2022locating}.

\subsection{Motivation: Prediction Drift in LVLMs}
\label{subsec:motivation}

We begin by examining how predictions evolve across layers in LVLMs. Using logit lens projections on LLaVA-1.5-7B, we track the probability of correct answer tokens at each layer for samples from POPE-Adversarial.

\begin{figure}[h]
\centering
\includegraphics[width=\linewidth]{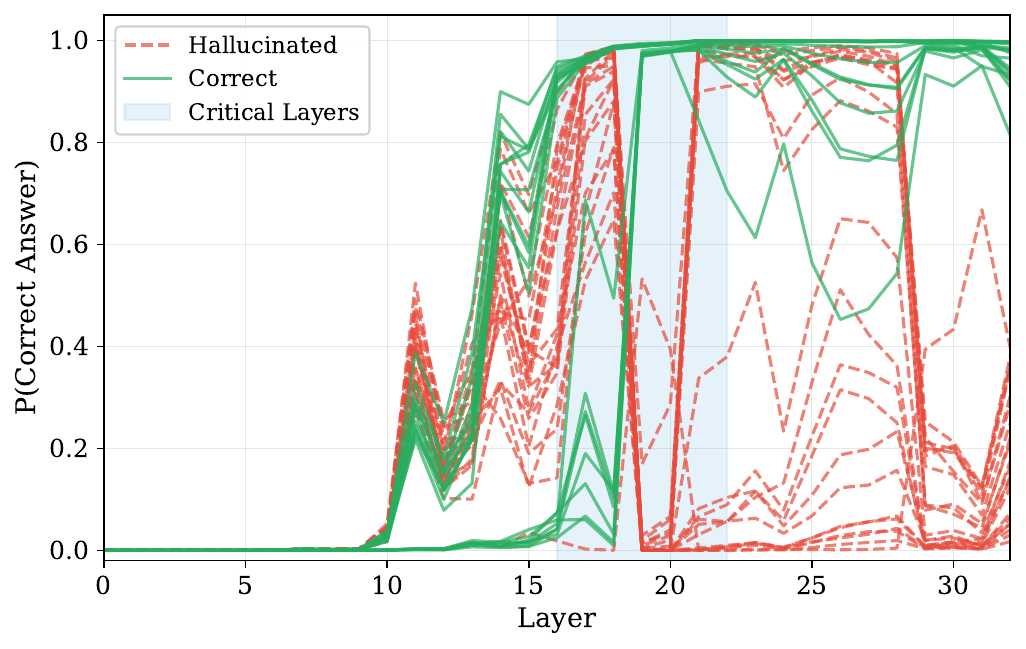}
\caption{P(Correct Answer) trajectories across layers for hallucinated (red, dashed) and correct (green, solid) samples. Correct samples maintain high probability throughout, while hallucinated samples drift to low probability in later layers. The shaded region indicates critical layers (16--22).}
\vspace{-4mm}
\label{fig:drift}

\end{figure}

Figure~\ref{fig:drift} reveals a striking pattern: for hallucinated samples, predictions drift from high to low P(Correct Answer) in later layers, while correct samples maintain stable high probability throughout. This observation raises a critical question: \emph{what causes this prediction drift?} We address this through mechanistic analysis in the following sections.

\subsection{Mechanistic Analysis}
\label{subsec:analysis}

To understand what causes the prediction drift observed in Figure~\ref{fig:drift}, we decompose the contributions of attention and FFN modules at each layer. We analyze LLaVA-1.5-7B on 50 samples from POPE-Adversarial.

\noindent\textbf{Contribution Analysis.}\quad
To measure each component's contribution, we use a differential logit lens approach. For attention at layer $l$:
\begin{equation}
    \Delta_{\mathrm{Attn}}^{(l)}(t) = \mathrm{LM}_{\mathrm{head}}(\tilde{\mathbf{h}}^{(l)})_{t} - \mathrm{LM}_{\mathrm{head}}(\mathbf{h}^{(l)})_{t}
\end{equation}
where $t$ is the target token. We compute $\Delta_{\mathrm{FFN}}^{(l)}$ analogously. This differential approach accounts for the nonlinearity of layer normalization.

\noindent\textbf{Correct-Direction Metric.}\quad
To enable comparison across samples with different ground truths, we define a \emph{correct-direction} metric:
\begin{equation}
    C^{(l)} = \Delta^{(l)}(t_{\mathrm{correct}}) - \Delta^{(l)}(t_{\mathrm{incorrect}})
\end{equation}
Under this metric, $C^{(l)} > 0$ indicates the component pushes toward the correct answer, while $C^{(l)} < 0$ indicates it pushes toward the wrong answer.

\begin{table}[h]
\centering
\small
\begin{tabular}{lcccc}
\toprule
\textbf{Prediction} & \textbf{Attn} & \textbf{FFN\textsubscript{total}} & \textbf{FFN\textsubscript{16-22}} & \textbf{FFN\textsubscript{L18}} \\
\midrule
Correct ($n$=40) & $+1.2$ & $+1.7$ & $\mathbf{+8.4}$ & $+6.0$ \\
Wrong ($n$=10) & $+0.8$ & $-2.0$ & $\mathbf{-3.5}$ & $-2.4$ \\
\bottomrule
\end{tabular}
\caption{Mean contributions toward correct answer (correct-direction metric). Values are summed across layers and averaged across samples. Positive values indicate pushing toward ground truth. FFN at layers 16--22 shows the largest difference between correct and wrong predictions.}
\label{tab:contribution}
\vspace{-4mm}
\end{table}

\noindent\textbf{OBS-1: Attention Aggregates Visual Evidence.}\quad
Attention contributions are positive and comparable for both correct ($+1.2$) and hallucinated ($+0.8$) samples (Table~\ref{tab:contribution}). This indicates that attention consistently aggregates visual features toward correct predictions across all samples.

\noindent\textbf{OBS-2: FFN Introduces Language Priors.}\quad
In contrast, FFN at layers 16--22 shows a striking difference: $+8.4$ for correct predictions and $-3.5$ for wrong predictions. For correct samples, FFN reinforces the prediction; for hallucinated samples, FFN actively pushes toward the wrong answer. We identify FFN as the source of language priors---when these priors conflict with visual evidence, they can override attention's correct predictions.

This directly explains the drift in Figure~\ref{fig:drift}: attention establishes correct predictions in intermediate layers, but language priors from FFN at layers 16--22 override the visual evidence, causing the prediction to drift toward incorrect outputs.

\subsection{FADE: FFN Attenuation for Decoding}
\label{subsec:fade}

Based on our analysis, we propose \textbf{FADE}, which attenuates FFN outputs at critical layers to reduce language-prior dominance:
\begin{equation}
    \mathbf{h}^{(l+1)} = \tilde{\mathbf{h}}^{(l)} + (1 - \alpha) \cdot \mathrm{FFN}^{(l)}(\tilde{\mathbf{h}}^{(l)})
\label{eq:fade}
\end{equation}
where $\tilde{\mathbf{h}}^{(l)}$ is the post-attention hidden state and $\alpha \in [0, 1]$ is the attenuation strength. The method is training-free, requires no additional parameters, and introduces negligible overhead. By reducing FFN contributions at selected critical layers, FADE suppresses language priors while preserving visual evidence aggregated by attention. We identify layers 16--22 as the critical band on LLaVA-1.5-7B and select task-specific intervention layers within this band; for other architectures, we transfer the band to proportionally equivalent mid-to-late layers, with full per-model configurations reported in Appendix~\ref{sec:appendix_hyperparams}.

%% file: sections/4_experiments_updated_ref.tex
\section{Experiments}
\label{sec:experiments}

\input{tables/pope_combined}

\subsection{Experimental Setup}
\label{subsec:exp_setup}

\paragraph{Models.}
We evaluate FADE on three representative LVLMs spanning diverse architectures:
\textbf{LLaVA-1.5-7B}~\cite{liu2024improved}, which uses a two-stage training with visual instruction tuning;
\textbf{mPLUG-Owl2-7B}~\cite{ye2024mplug}, which employs modality-adaptive modules for vision-language alignment; and
\textbf{InstructBLIP-7B}~\cite{dai2023instructblip}, which introduces instruction-aware visual feature extraction via Q-Former.
This selection covers the major LVLM design paradigms and enables comprehensive evaluation of our method's generalizability.

\paragraph{Benchmarks.}
We adopt three widely-used benchmarks:
\textbf{POPE}~\cite{li2023evaluating} probes object hallucination via binary (Yes/No) questions across three sampling strategies (Random, Popular, Adversarial) on MSCOCO, A-OKVQA, and GQA;
\textbf{CHAIR}~\cite{rohrbach2018object} measures hallucination in image captioning, where CHAIR$_S$ and CHAIR$_I$ denote sentence-level and instance-level hallucination rates (lower is better) and Recall measures coverage (higher is better);
\textbf{MME}~\cite{fu2025mme} evaluates perception and cognition across 14 subtasks, and we report perception scores across its ten perception subtasks.

\paragraph{Baselines.}
We compare against representative training-free methods from each category:
\textbf{PAI}~\cite{liu2024paying} amplifies attention on image tokens;
\textbf{VCD}~\cite{leng2024mitigating} contrasts outputs from original and distorted images;
\textbf{DAMO}~\cite{wang2025damo} applies momentum-based activation stabilization;
\textbf{VISTA}~\cite{li2025hidden} steers representations using pre-computed visual vectors; and
\textbf{DCLA}~\cite{tang2025mitigating} enforces inter-layer consistency via layer aggregation.
All baselines use official implementations with recommended hyperparameters. VISTA relies on a model-specific visual steering vector computation that is only officially supported on its three released architectures; we therefore evaluate it on LLaVA-1.5, mPLUG-Owl2, and InstructBLIP, and substitute DCLA on the advanced models (InternVL3-8B, Qwen2.5/3-VL) as a representative baseline from the layer-aggregation family.

\paragraph{Implementation.}
All experiments use greedy decoding on 8 NVIDIA H100 80GB GPUs.
FADE attenuates FFN outputs at task-specific critical layers selected from the mid-to-late critical-layer band (Section~\ref{subsec:fade}). On LLaVA-1.5-7B, we use $\alpha=0.6$ at layer 18 based on the ablation in Section~\ref{subsec:ablation}; per-model layer indices for mPLUG-Owl2, InstructBLIP, InternVL3-8B, and the Qwen-VL series are obtained by mapping the LLaVA critical band to the proportionally-equivalent mid-to-late layers, with all hyperparameters and baseline configurations detailed in Appendix~\ref{sec:appendix_settings} and~\ref{sec:appendix_hyperparams}.

\input{tables/chair_combined}
\input{tables/mme_combined}
\subsection{Main Results}
\label{subsec:main_results}

\subsubsection{Results on POPE}

Table~\ref{tab:pope_combined} presents results under random, popular, and adversarial settings.
FADE is strongest or tied under the challenging GQA adversarial setting on LLaVA-1.5 and mPLUG-Owl2, achieving 82.5\% and 79.0\% F1, respectively.
On LLaVA-1.5, FADE surpasses VCD by 5.0\% F1 and DAMO by 1.7\% F1 on the GQA adversarial subset.
Notably, VCD shows limited generalization on LLaVA-1.5 GQA (72.0\% accuracy under adversarial), likely because its contrastive decoding with noisy images disrupts fine-grained spatial reasoning required for GQA's scene graph questions. DAMO and VISTA improve over greedy decoding but exhibit inconsistent behavior---DAMO gains on A-OKVQA but plateaus on GQA, while VISTA shows marginal improvements that do not consistently exceed the baseline across all settings. Results on LLaVA-v1.5-13B, reported in Appendix~\ref{sec:appendix_13b}, show that FADE maintains its lead over all baselines at larger scales, while VCD, DAMO, and VISTA all degrade below greedy decoding.

\subsubsection{Results on CHAIR}
Table~\ref{tab:chair_combined} reports image captioning results.
Among existing methods, we observe a clear accuracy-coverage trade-off: VISTA achieves the lowest CHAIR$_S$ (19.2\%) on LLaVA-1.5 but sacrifices Recall significantly (62.6\% vs.\ 80.6\% for greedy), indicating over-aggressive suppression of generation.
Conversely, VCD and DAMO increase hallucination rates on most models---VCD raises CHAIR$_S$ from 49.8\% to 58.6\% on LLaVA-1.5, suggesting that their uniform intervention strategies disrupt fluent generation.

Relative to greedy decoding, FADE reduces both CHAIR$_S$ and CHAIR$_I$ across all three models.
On mPLUG-Owl2, FADE achieves the lowest CHAIR$_S$ (55.0\%) among all methods.
On InstructBLIP, FADE substantially reduces instance-level hallucination to CHAIR$_I$=14.0\%, compared to 37.9\% for PAI and 38.5\% for greedy, while maintaining competitive Recall (72.9\%). All methods are evaluated under identical decoding configurations and on the same caption pool, ensuring a fair comparison (full settings in Appendix~\ref{sec:appendix_settings}).
This pronounced effect suggests that FFN attenuation is particularly well-matched to Q-Former-based visual encoders, where pooled visual queries tend to leave more residual capacity for FFN-stored priors to dominate.

\subsubsection{Results on MME}

Table~\ref{tab:mme_full} reports MME perception scores across 10 subtasks.
On LLaVA-1.5, FADE achieves 1519.0 total perception score, improving over greedy decoding (1505.7) by +13.3 points and outperforming all baselines including PAI (1508.9).
The improvement is particularly notable on counting (+5.0 over greedy) and celebrity recognition (+1.7), subtasks that require precise object grounding.
Interestingly, different methods show architecture-dependent behavior: PAI improves LLaVA-1.5 (+3.2) but slightly degrades mPLUG-Owl2 ($-$15.6), while DAMO gains substantially on mPLUG-Owl2's counting subtask (+10.0) but loses on LLaVA-1.5 ($-$6.7).
This architecture sensitivity suggests that attention-based and contrastive methods may interact differently with each model's vision-language alignment mechanism.
FADE's FFN-level intervention provides a more architecture-agnostic approach by targeting the representation drift phenomenon that is common across transformer-based LVLMs.

\subsubsection{Results on MMHal-Bench}

We further evaluate on MMHal-Bench, where GPT-4 judges open-ended responses across eight categories, testing whether mitigation methods generalize beyond binary Yes/No questions to free-form generation~\cite{sun2024aligning}. Table~\ref{tab:mmhal} shows that FADE achieves the highest overall GPT-4 judged score (2.09 vs.\ 2.05 for greedy, 1.83 for PAI, 1.92 for VCD), indicating that FFN attenuation preserves---rather than degrades---generation quality in open-ended settings.
\input{tables/mmhal}

\subsubsection{Generalization to Advanced Architectures}
\label{subsubsec:sota_generalization}

To evaluate architecture-agnostic robustness, we extend FADE to next-generation models: \textbf{InternVL3-8B} (Table~\ref{tab:internvl3_full}) and the \textbf{Qwen-VL series} (Qwen2.5-VL-7B, Table~\ref{tab:qwen2.5_results}; Qwen3-VL-8B, Table~\ref{tab:qwen3_results}). On InternVL3-8B, FADE attains the top MMBench$^{\text{EN}}$ score (\textbf{69.24\%}, +3.1\% over greedy) and the highest MME Perception (\textbf{1734.6}), while remaining competitive on the adversarial POPE split (88.2 F1, matching PAI); notably, on the open-ended CHAIR task all training-free interventions fail to beat greedy (29.2), suggesting that stronger models possess highly optimized internal language priors that aggressive modifications can easily disrupt. On Qwen2.5-VL FADE matches the highest MME (\textbf{1694.1}) and ties greedy for the second-best CHAIR$_S$ (36.6), with POPE-Adv (86.8) marginally improving over greedy (86.7); on Qwen3-VL it preserves the peak MMBench score (\textbf{86.5}), reaches the second-best POPE-Adv (\underline{88.2}, behind DAMO's 88.4), and reduces CHAIR$_S$ from 57.4 to 55.8. Across all three architectures, FADE delivers the most balanced trade-off, whereas alternatives such as DAMO on Qwen3-VL push HalBench to 58.3 but spike CHAIR$_S$ to 61.0, and VCD on Qwen3-VL reaches the lowest CHAIR$_S$ (26.6) at the cost of POPE-Adv (87.4, the lowest among all methods)---supporting FADE as a stable, architecture-agnostic intervention. A sensitivity sweep over the attenuation strength $\alpha$ on Qwen3-VL-8B is reported in Appendix~\ref{subsec:app_alpha_sweep}, showing that the gains are stable across $\alpha \in [0.3, 0.8]$ and not the product of cherry-picked tuning.

\begin{table}[!htbp]
\centering
\small
\setlength{\tabcolsep}{2pt}
\caption{Performance on InternVL3-8B. POPE is reported as Random/Popular/Adversarial F1 and their Average. Best \textbf{bold}, second best \underline{underlined}.}
\label{tab:internvl3_full}
\resizebox{\columnwidth}{!}{%
\begin{tabular}{lcccccccccc}
\toprule
\multirow{2}{*}{\textbf{Method}} & \multicolumn{4}{c}{\textbf{POPE} (F1)$\uparrow$} & \multicolumn{2}{c}{\textbf{CHAIR}$\downarrow$} & \multirow{2}{*}{\textbf{HalBench}$\uparrow$} & \multicolumn{2}{c}{\textbf{MME}$\uparrow$} & \multirow{2}{*}{\textbf{MMB}$^{\text{EN}}$$\uparrow$} \\
\cmidrule(lr){2-5} \cmidrule(lr){6-7} \cmidrule(lr){9-10}
& Rand & Pop & Adv & Avg & $C_S$ & $C_I$ & & Overall & Percep. & \\
\midrule
Greedy & 93.4 & \underline{90.9} & \underline{88.5} & 90.9 & \textbf{29.2} & \underline{7.4} & \underline{50.32} & 2359.8 & 1727.7 & 66.15 \\
PAI \textsubscript{\textit{ECCV'24}} & \underline{93.7} & \underline{90.9} & 88.2 & \underline{91.0} & \underline{29.6} & 7.6 & 50.06 & \textbf{2382.9} & 1723.6 & \underline{67.61} \\
VCD \textsubscript{\textit{CVPR'24}} & \textbf{93.8} & \textbf{91.3} & 88.1 & \textbf{91.1} & 31.4 & 8.3 & 47.50 & 2337.8 & 1713.9 & 64.95 \\
DCLA \textsubscript{\textit{arXiv'25}} & 93.2 & 90.7 & 88.4 & 90.8 & \underline{29.6} & \textbf{7.3} & 49.23 & \underline{2368.1} & \underline{1727.8} & 67.10 \\
DAMO \textsubscript{\textit{ICLR'25}} & 92.4 & 90.6 & \textbf{88.9} & 90.6 & 30.0 & \underline{7.4} & \textbf{51.94} & 2368.0 & 1713.3 & 66.92 \\
\cellcolor[rgb]{.9,.9,.9}\textbf{FADE} & \cellcolor[rgb]{.9,.9,.9}\underline{93.7} & \cellcolor[rgb]{.9,.9,.9}\underline{90.9} & \cellcolor[rgb]{.9,.9,.9}88.2 & \cellcolor[rgb]{.9,.9,.9}90.9 & \cellcolor[rgb]{.9,.9,.9}31.0 & \cellcolor[rgb]{.9,.9,.9}7.7 & \cellcolor[rgb]{.9,.9,.9}48.34 & \cellcolor[rgb]{.9,.9,.9}2366.7 & \cellcolor[rgb]{.9,.9,.9}\textbf{1734.6} & \cellcolor[rgb]{.9,.9,.9}\textbf{69.24} \\
\bottomrule
\end{tabular}%
}
\vspace{-3mm}
\end{table}

\begin{table}[!htbp]
\centering
\small
\setlength{\tabcolsep}{2pt}
\caption{Performance on Qwen2.5-VL-7B-Instruct. POPE is reported as Random/Popular/Adversarial F1 and their Average. Best \textbf{bold}, second best \underline{underlined}.}
\label{tab:qwen2.5_results}
\resizebox{\columnwidth}{!}{%
\begin{tabular}{lccccccccc}
\toprule
\multirow{2}{*}{\textbf{Method}} & \multicolumn{4}{c}{\textbf{POPE} (F1)$\uparrow$} & \multirow{2}{*}{\textbf{MME}$\uparrow$} & \multirow{2}{*}{\textbf{HalBench}$\uparrow$} & \multirow{2}{*}{\textbf{MMB}$\uparrow$} & \multirow{2}{*}{\textbf{CHAIR}$_S$$\downarrow$} & \multirow{2}{*}{\textbf{CHAIR}$_I$$\downarrow$} \\
\cmidrule(lr){2-5}
& Rand & Pop & Adv & Avg & & & & & \\
\midrule
Greedy & 88.6 & 87.8 & 86.7 & 87.7 & \underline{1690.4} & \underline{55.9} & 84.8 & \underline{36.6} & 9.5 \\
PAI & \underline{89.3} & \underline{88.4} & \underline{87.2} & \underline{88.3} & 1662.4 & 55.7 & \underline{84.9} & 37.2 & 8.4 \\
DCLA & 89.2 & \underline{88.4} & 87.1 & 88.2 & 1690.1 & 54.9 & 84.1 & 37.4 & \underline{7.6} \\
DAMO & 88.3 & 87.6 & 86.5 & 87.4 & 1685.4 & 55.2 & 84.1 & 37.6 & \textbf{7.1} \\
VCD & \textbf{89.5} & \textbf{88.7} & \textbf{87.6} & \textbf{88.6} & 1689.4 & \textbf{56.2} & \textbf{85.2} & \textbf{32.4} & 7.8 \\
\cellcolor[rgb]{.9,.9,.9}\textbf{FADE} & \cellcolor[rgb]{.9,.9,.9}88.6 & \cellcolor[rgb]{.9,.9,.9}87.9 & \cellcolor[rgb]{.9,.9,.9}86.8 & \cellcolor[rgb]{.9,.9,.9}87.8 & \cellcolor[rgb]{.9,.9,.9}\textbf{1694.1} & \cellcolor[rgb]{.9,.9,.9}55.2 & \cellcolor[rgb]{.9,.9,.9}\underline{84.9} & \cellcolor[rgb]{.9,.9,.9}\underline{36.6} & \cellcolor[rgb]{.9,.9,.9}9.5 \\
\bottomrule
\end{tabular}%
}
\vspace{-3mm}
\end{table}

\begin{table}[!htbp]
\centering
\small
\setlength{\tabcolsep}{2pt}
\caption{Performance on Qwen3-VL-8B-Instruct. POPE is reported as Random/Popular/Adversarial F1 and their Average. Best \textbf{bold}, second best \underline{underlined}.}
\label{tab:qwen3_results}
\resizebox{\columnwidth}{!}{%
\begin{tabular}{lccccccccc}
\toprule
\multirow{2}{*}{\textbf{Method}} & \multicolumn{4}{c}{\textbf{POPE} (F1)$\uparrow$} & \multirow{2}{*}{\textbf{MME}$\uparrow$} & \multirow{2}{*}{\textbf{HalBench}$\uparrow$} & \multirow{2}{*}{\textbf{MMB}$\uparrow$} & \multirow{2}{*}{\textbf{CHAIR}$_S$$\downarrow$} & \multirow{2}{*}{\textbf{CHAIR}$_I$$\downarrow$} \\
\cmidrule(lr){2-5}
& Rand & Pop & Adv & Avg & & & & & \\
\midrule
Greedy & 92.2 & \underline{89.5} & 88.0 & 89.9 & 1745.9 & 56.8 & \textbf{86.5} & 57.4 & 10.3 \\
PAI & \underline{92.6} & 89.6 & 87.9 & \underline{90.0} & 1735.8 & \underline{57.8} & 85.7 & \underline{55.4} & 9.8 \\
DCLA & 92.2 & \underline{89.5} & 88.0 & 89.9 & \textbf{1751.4} & 56.1 & \underline{86.4} & 56.4 & 10.0 \\
DAMO & 92.2 & \textbf{89.7} & \textbf{88.4} & \textbf{90.1} & \underline{1749.8} & \textbf{58.3} & \underline{86.4} & 61.0 & \underline{9.5} \\
VCD & \textbf{92.6} & 89.2 & 87.4 & 89.7 & 1743.9 & 55.3 & 85.7 & \textbf{26.6} & \textbf{8.0} \\
\cellcolor[rgb]{.9,.9,.9}\textbf{FADE} & \cellcolor[rgb]{.9,.9,.9}92.2 & \cellcolor[rgb]{.9,.9,.9}\underline{89.5} & \cellcolor[rgb]{.9,.9,.9}\underline{88.2} & \cellcolor[rgb]{.9,.9,.9}\underline{90.0} & \cellcolor[rgb]{.9,.9,.9}1746.2 & \cellcolor[rgb]{.9,.9,.9}57.2 & \cellcolor[rgb]{.9,.9,.9}\textbf{86.5} & \cellcolor[rgb]{.9,.9,.9}55.8 & \cellcolor[rgb]{.9,.9,.9}9.9 \\
\bottomrule
\end{tabular}%
}
\vspace{-3mm}
\end{table}

\subsection{Efficiency Study}
\label{subsec:efficiency}

We analyze FADE's computational efficiency compared to existing methods.
Table~\ref{tab:efficiency} compares inference efficiency.
FADE adds only 3\% latency overhead compared to greedy decoding (122ms vs 118ms), while achieving substantial speedups over all comparison methods: 19\% faster than DAMO, 34\% faster than PAI, 57\% faster than VCD, and 73\% faster than VISTA.
VCD requires a second forward pass with distorted images, resulting in 2.4$\times$ total latency.
VISTA incurs the highest overhead (3.9$\times$) due to steering vector computation during inference.
FADE's efficiency stems from: (1) FFN attenuation requiring only element-wise scaling at a single layer, not additional forward passes; and (2) no memory overhead (14.5 GB, identical to greedy decoding).
This single-pass design is complementary to recent efforts on efficient multimodal and LLM reasoning that compress chain-of-thought traces~\cite{zhang2026chain}, address late-stage fragility in reasoning chains~\cite{zhang2025not}, or adaptively allocate compute via coarse-to-fine refinement~\cite{zhang2026not}, suggesting that FADE can be combined with such orthogonal acceleration techniques.

\input{tables/efficiency}

\subsection{Case Study}
\label{subsec:case_study}
\begin{figure}[h]
\centering
\includegraphics[width=0.95\columnwidth]{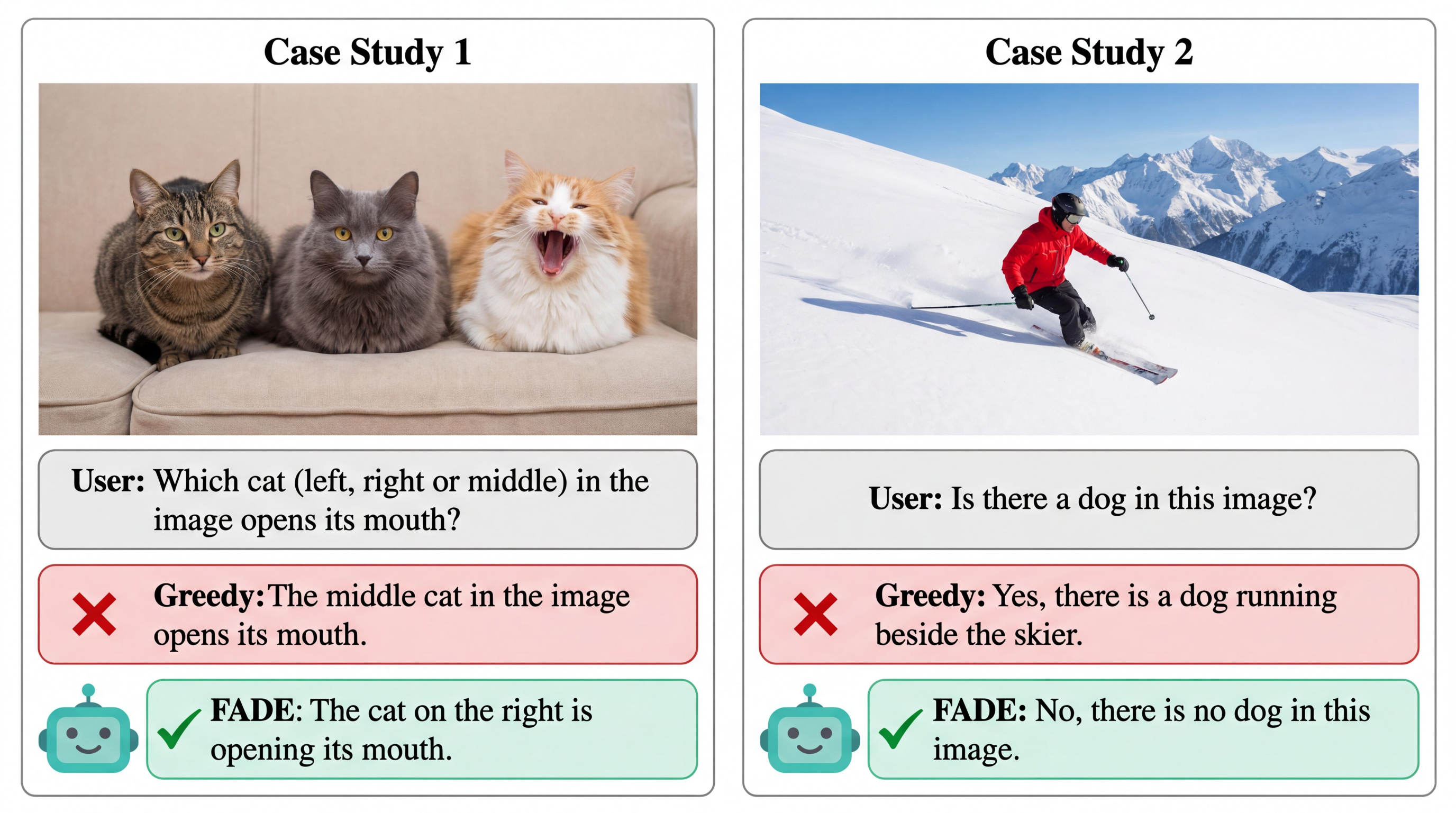}
\caption{Qualitative comparison of hallucination correction. Case Study 1: Greedy decoding incorrectly identifies which cat opens its mouth, while FADE provides the correct answer. Case Study 2: Greedy decoding hallucinates a non-existent dog in the skiing scene, while FADE correctly denies its presence.}
\label{fig:case_studies}
\vspace{-4mm}
\end{figure}

Figure~\ref{fig:case_studies} illustrates how FFN attenuation mitigates language prior dominance: FADE resolves spatial reasoning errors (Case 1) and suppresses non-existent object hallucinations (Case 2), yielding more visually grounded responses.

\subsection{Ablation Study}
\label{subsec:ablation}

\begin{figure}[h]
\centering
\begin{subfigure}[b]{0.48\columnwidth}
\centering
\includegraphics[width=\columnwidth]{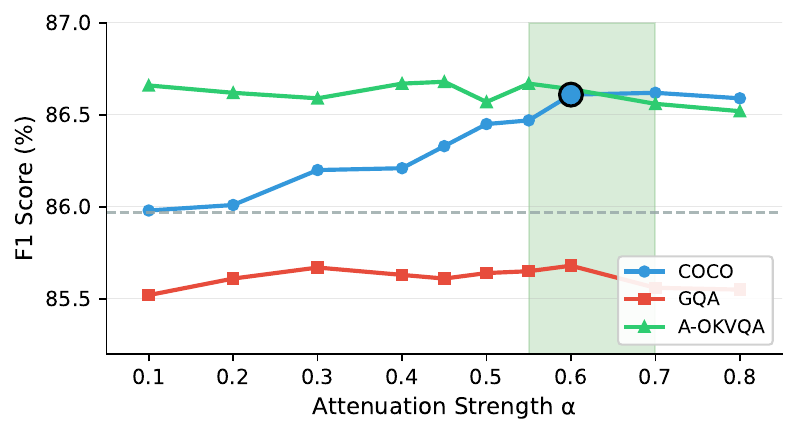}
\caption{LLaVA-1.5: Strength}
\label{fig:ablation_strength}
\end{subfigure}
\hfill
\begin{subfigure}[b]{0.48\columnwidth}
\centering
\includegraphics[width=\columnwidth]{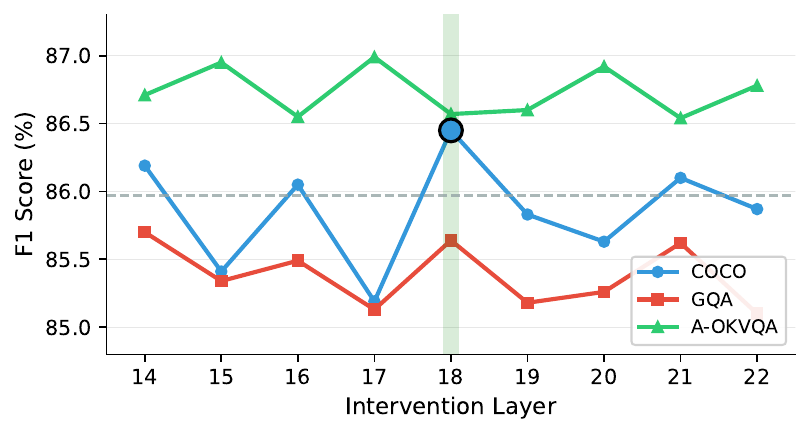}
\caption{LLaVA-1.5: Layer}
\label{fig:ablation_layer}
\end{subfigure}
\\
\begin{subfigure}[b]{0.48\columnwidth}
\centering
\includegraphics[width=\columnwidth]{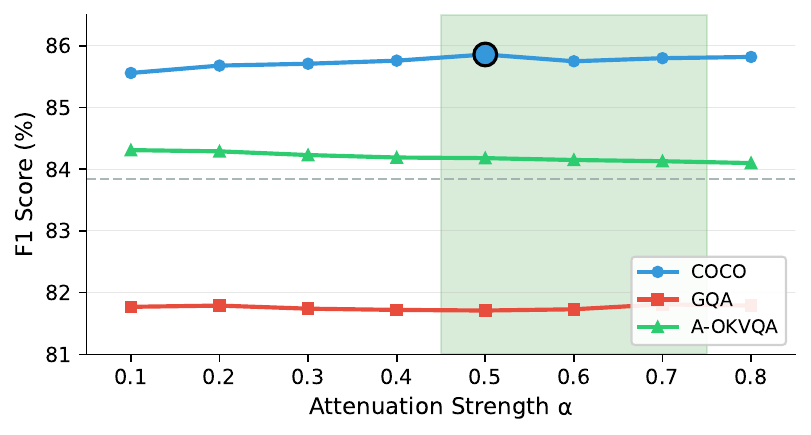}
\caption{mPLUG-Owl2: Strength}
\label{fig:mplug_ablation_strength}
\end{subfigure}
\hfill
\begin{subfigure}[b]{0.48\columnwidth}
\centering
\includegraphics[width=\columnwidth]{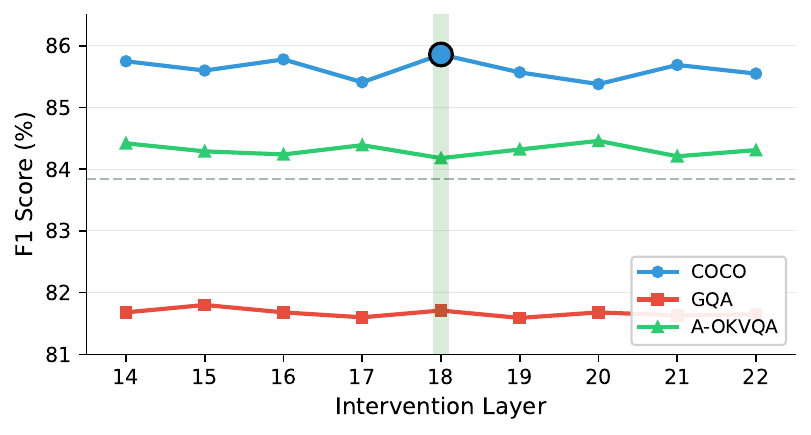}
\caption{mPLUG-Owl2: Layer}
\label{fig:mplug_ablation_layer}
\end{subfigure}
\caption{Ablation on POPE. (a)(c) Strength sensitivity: optimal range $[0.5, 0.7]$. (b)(d) Layer sensitivity: Layer~18 provides the best or near-best trade-off on LLaVA-1.5 and mPLUG-Owl2. Shaded regions indicate recommended hyperparameter ranges.}
\label{fig:ablation_combined}
\vspace{-4mm}
\end{figure}

We conduct ablations on POPE to analyze hyperparameter sensitivity across different models.

\paragraph{Strength and Layer.}
Varying $\alpha \in [0.1, 0.8]$ (Figure~\ref{fig:ablation_strength},~\ref{fig:mplug_ablation_strength}) yields optimal F1 at $\alpha{=}0.6$ on LLaVA-1.5 (variation within 0.3\% across $[0.55, 0.7]$) and $\alpha{=}0.5$--$0.7$ on mPLUG-Owl2, showing consistent low sensitivity across architectures. Sweeping intervention layers 14--22 (Figure~\ref{fig:ablation_layer},~\ref{fig:mplug_ablation_layer}) identifies Layer~18 as optimal on both models, matching our analysis that mid-to-late layers exhibit the highest directional drift; mPLUG-Owl2 shows mild dataset-dependent variation, with A-OKVQA preferring layers 14/20 and COCO/GQA favoring 18.

\paragraph{Task-Specific Tuning.} Optimal hyperparameters vary by task (Appendix~\ref{sec:appendix_ablation}): discriminative POPE prefers $\alpha{=}0.6$, generative CHAIR benefits from stronger attenuation ($\alpha{=}1.0$) at later layers (L20), while MME's diverse reasoning requires gentler intervention ($\alpha{=}0.02$).

%% file: tables/pope_combined.tex
\begin{table*}[!htbp]
\centering
\renewcommand{\arraystretch}{0.95}
\setlength{\tabcolsep}{2.5pt}
\caption{POPE benchmark results across three VLMs. We evaluate across three sampling strategies (Random, Popular, Adversarial) and three datasets (MSCOCO, A-OKVQA, GQA). Best results are in \textbf{bold}, second best are \underline{underlined}.}
\vspace{-3mm}
\label{tab:pope_combined}
\resizebox{\textwidth}{!}{%
\begin{tabular}{ll|cccccc|cccccc|cccccc}
\toprule
& & \multicolumn{6}{c|}{\textbf{LLaVA-1.5-7B}} & \multicolumn{6}{c|}{\textbf{mPLUG-Owl2-7B}} & \multicolumn{6}{c}{\textbf{InstructBLIP-7B}} \\
\cmidrule(lr){3-8} \cmidrule(lr){9-14} \cmidrule(lr){15-20}
& & \multicolumn{2}{c|}{MSCOCO} & \multicolumn{2}{c|}{A-OKVQA} & \multicolumn{2}{c|}{GQA} & \multicolumn{2}{c|}{MSCOCO} & \multicolumn{2}{c|}{A-OKVQA} & \multicolumn{2}{c|}{GQA} & \multicolumn{2}{c|}{MSCOCO} & \multicolumn{2}{c|}{A-OKVQA} & \multicolumn{2}{c}{GQA} \\
\cmidrule(lr){3-8} \cmidrule(lr){9-14} \cmidrule(lr){15-20}
\textbf{Setting} & \textbf{Method} & \multicolumn{2}{c|}{Acc \hspace{1pt} F1} & \multicolumn{2}{c|}{Acc \hspace{1pt} F1} & \multicolumn{2}{c|}{Acc \hspace{1pt} F1} & \multicolumn{2}{c|}{Acc \hspace{1pt} F1} & \multicolumn{2}{c|}{Acc \hspace{1pt} F1} & \multicolumn{2}{c|}{Acc \hspace{1pt} F1} & \multicolumn{2}{c|}{Acc \hspace{1pt} F1} & \multicolumn{2}{c|}{Acc \hspace{1pt} F1} & \multicolumn{2}{c}{Acc \hspace{1pt} F1} \\
\midrule
\multirow{6}{*}{\rotatebox{90}{\textbf{Random}}}
& Greedy & 88.5 & 87.3 & 91.0 & 90.7 & 89.3 & 89.0 & \underline{88.2} & 87.4 & \underline{88.5} & \underline{88.4} & \underline{86.9} & \underline{86.1} & 87.2 & 85.8 & \underline{88.6} & \underline{88.4} & \underline{87.4} & \underline{87.1} \\
& PAI \textsubscript{\textit{ECCV'24}} & 88.5 & 87.4 & 91.0 & 90.7 & 89.2 & 88.8 & \textbf{88.5} & \underline{87.8} & 88.4 & 88.2 & 86.6 & 85.7 & 69.0 & 73.1 & 66.2 & 71.8 & 64.4 & 70.6 \\
& VCD \textsubscript{\textit{CVPR'24}} & \textbf{89.9} & \textbf{90.0} & 88.5 & 89.4 & 88.1 & 89.0 & 87.9 & \textbf{88.1} & 85.0 & 85.9 & 85.2 & 85.6 & \textbf{89.1} & \textbf{88.6} & 87.1 & 87.7 & 86.3 & 87.0 \\
& DAMO \textsubscript{\textit{ICLR'25}} & 86.4 & 84.6 & 89.0 & 88.2 & 86.4 & 85.2 & 87.9 & 87.0 & \textbf{88.6} & \underline{88.4} & 86.2 & 85.2 & \underline{88.2} & \underline{88.1} & 83.7 & 85.2 & 84.4 & 85.8 \\
& VISTA \textsubscript{\textit{ICML'25}} & 88.8 & 87.8 & \textbf{91.4} & \textbf{91.3} & \textbf{90.0} & \textbf{89.8} & \underline{88.2} & 87.4 & \underline{88.5} & \underline{88.4} & 86.7 & 85.8 & 88.0 & 86.9 & 88.5 & \textbf{88.6} & \textbf{87.7} & \textbf{87.8} \\
& \cellcolor[rgb]{.9,.9,.9}\textbf{FADE} & \cellcolor[rgb]{.9,.9,.9}\underline{89.2} & \cellcolor[rgb]{.9,.9,.9}\underline{88.3} & \cellcolor[rgb]{.9,.9,.9}\underline{91.2} & \cellcolor[rgb]{.9,.9,.9}\underline{91.1} & \cellcolor[rgb]{.9,.9,.9}\underline{89.8} & \cellcolor[rgb]{.9,.9,.9}\underline{89.7} & \cellcolor[rgb]{.9,.9,.9}\textbf{88.5} & \cellcolor[rgb]{.9,.9,.9}\underline{87.8} & \cellcolor[rgb]{.9,.9,.9}\textbf{88.6} & \cellcolor[rgb]{.9,.9,.9}\textbf{88.5} & \cellcolor[rgb]{.9,.9,.9}\textbf{87.0} & \cellcolor[rgb]{.9,.9,.9}\textbf{86.2} & \cellcolor[rgb]{.9,.9,.9}86.9 & \cellcolor[rgb]{.9,.9,.9}85.5 & \cellcolor[rgb]{.9,.9,.9}\textbf{88.9} & \cellcolor[rgb]{.9,.9,.9}\textbf{88.6} & \cellcolor[rgb]{.9,.9,.9}\underline{87.4} & \cellcolor[rgb]{.9,.9,.9}87.0 \\
\midrule
\multirow{6}{*}{\rotatebox{90}{\textbf{Popular}}}
& Greedy & 87.2 & 86.1 & \underline{87.6} & \underline{87.6} & \textbf{84.5} & \underline{84.7} & \underline{86.2} & 85.5 & 84.6 & 85.0 & \underline{80.0} & \textbf{80.2} & 84.8 & 83.6 & \underline{81.3} & \underline{82.2} & \underline{77.1} & 78.8 \\
& PAI \textsubscript{\textit{ECCV'24}} & \underline{87.4} & 86.3 & \textbf{87.8} & \textbf{87.8} & \textbf{84.5} & \underline{84.7} & \textbf{86.4} & \textbf{85.9} & 84.6 & 85.0 & 79.9 & 80.0 & 65.0 & 70.6 & 59.2 & 66.8 & 55.7 & 65.9 \\
& VCD \textsubscript{\textit{CVPR'24}} & 86.6 & \textbf{87.1} & 82.3 & 84.5 & 76.6 & 80.4 & 83.6 & 84.4 & 81.2 & 82.9 & 77.5 & 79.4 & \underline{85.2} & \textbf{85.2} & 79.9 & 82.0 & \underline{77.1} & \textbf{80.0} \\
& DAMO \textsubscript{\textit{ICLR'25}} & 85.4 & 83.7 & 87.2 & 86.6 & \underline{84.2} & 83.2 & \textbf{86.4} & \underline{85.6} & \textbf{84.9} & \textbf{85.2} & \textbf{80.2} & 80.0 & 82.4 & 83.3 & 74.7 & 78.8 & 71.5 & 76.8 \\
& VISTA \textsubscript{\textit{ICML'25}} & \underline{87.4} & 86.5 & 86.6 & 87.1 & 83.2 & 84.1 & \underline{86.2} & \underline{85.6} & \underline{84.7} & \underline{85.1} & \underline{80.0} & \underline{80.1} & \textbf{85.5} & \underline{84.6} & 80.3 & 81.9 & 75.8 & 78.5 \\
& \cellcolor[rgb]{.9,.9,.9}\textbf{FADE} & \cellcolor[rgb]{.9,.9,.9}\textbf{87.7} & \cellcolor[rgb]{.9,.9,.9}\underline{86.9} & \cellcolor[rgb]{.9,.9,.9}87.0 & \cellcolor[rgb]{.9,.9,.9}87.4 & \cellcolor[rgb]{.9,.9,.9}84.1 & \cellcolor[rgb]{.9,.9,.9}\textbf{84.8} & \cellcolor[rgb]{.9,.9,.9}\textbf{86.4} & \cellcolor[rgb]{.9,.9,.9}\textbf{85.9} & \cellcolor[rgb]{.9,.9,.9}84.5 & \cellcolor[rgb]{.9,.9,.9}\underline{85.1} & \cellcolor[rgb]{.9,.9,.9}79.9 & \cellcolor[rgb]{.9,.9,.9}\textbf{80.2} & \cellcolor[rgb]{.9,.9,.9}84.7 & \cellcolor[rgb]{.9,.9,.9}83.4 & \cellcolor[rgb]{.9,.9,.9}\textbf{81.9} & \cellcolor[rgb]{.9,.9,.9}\textbf{82.7} & \cellcolor[rgb]{.9,.9,.9}\textbf{77.8} & \cellcolor[rgb]{.9,.9,.9}\underline{79.3} \\
\midrule
\multirow{6}{*}{\rotatebox{90}{\textbf{Adversarial}}}
& Greedy & 85.1 & 84.2 & 80.4 & 81.8 & \textbf{81.5} & \underline{82.3} & \underline{84.0} & \underline{83.6} & 77.6 & \underline{79.7} & \underline{78.5} & \textbf{79.0} & \textbf{83.0} & \underline{82.0} & \underline{74.6} & \underline{77.3} & \underline{75.1} & \underline{77.3} \\
& PAI \textsubscript{\textit{ECCV'24}} & \textbf{85.3} & 84.3 & \underline{80.7} & \underline{81.9} & \textbf{81.5} & 82.2 & \textbf{84.2} & \textbf{83.9} & \underline{77.7} & \underline{79.7} & 78.4 & 78.8 & 62.8 & 69.3 & 55.7 & 63.0 & 55.4 & 65.7 \\
& VCD \textsubscript{\textit{CVPR'24}} & 81.2 & 82.7 & 73.3 & 78.4 & 72.0 & 77.5 & 79.9 & 81.6 & 73.7 & 77.9 & 75.0 & 77.9 & 82.0 & \textbf{82.5} & 72.0 & 76.6 & 72.9 & 77.2 \\
& DAMO \textsubscript{\textit{ICLR'25}} & 83.9 & 82.3 & \textbf{82.1} & \textbf{82.2} & \underline{81.4} & 80.8 & \textbf{84.2} & \underline{83.6} & \textbf{78.2} & \textbf{80.0} & \textbf{78.6} & 78.7 & 79.4 & 81.0 & 67.3 & 74.2 & 68.1 & 74.7 \\
& VISTA \textsubscript{\textit{ICML'25}} & \underline{85.2} & \underline{84.5} & 79.2 & 81.2 & 80.4 & 81.9 & \underline{84.0} & \underline{83.6} & \underline{77.7} & \underline{79.7} & \underline{78.5} & \underline{78.9} & \textbf{83.0} & \textbf{82.5} & 73.3 & 76.9 & 73.6 & 76.9 \\
& \cellcolor[rgb]{.9,.9,.9}\textbf{FADE} & \cellcolor[rgb]{.9,.9,.9}\underline{85.2} & \cellcolor[rgb]{.9,.9,.9}\textbf{84.6} & \cellcolor[rgb]{.9,.9,.9}79.5 & \cellcolor[rgb]{.9,.9,.9}81.4 & \cellcolor[rgb]{.9,.9,.9}81.1 & \cellcolor[rgb]{.9,.9,.9}\textbf{82.5} & \cellcolor[rgb]{.9,.9,.9}\textbf{84.2} & \cellcolor[rgb]{.9,.9,.9}\textbf{83.9} & \cellcolor[rgb]{.9,.9,.9}77.4 & \cellcolor[rgb]{.9,.9,.9}\underline{79.7} & \cellcolor[rgb]{.9,.9,.9}78.3 & \cellcolor[rgb]{.9,.9,.9}\textbf{79.0} & \cellcolor[rgb]{.9,.9,.9}\underline{82.9} & \cellcolor[rgb]{.9,.9,.9}81.9 & \cellcolor[rgb]{.9,.9,.9}\textbf{75.3} & \cellcolor[rgb]{.9,.9,.9}\textbf{77.8} & \cellcolor[rgb]{.9,.9,.9}\textbf{75.6} & \cellcolor[rgb]{.9,.9,.9}\textbf{77.6} \\
\bottomrule
\end{tabular}
}
\vspace{-4mm}
\end{table*}

%% file: tables/chair_combined.tex
\begin{table*}[t]
\centering
\scriptsize
\renewcommand{\arraystretch}{0.9}
\setlength{\tabcolsep}{3pt}
\caption{CHAIR benchmark results across three VLMs. C$_S$/C$_I$: sentence/instance-level hallucination rates (lower is better). Rec: recall (higher is better). Best results are in \textbf{bold}, second best are \underline{underlined}.}
\label{tab:chair_combined}
\resizebox{\textwidth}{!}{%
\begin{tabular}{l|cccc|cccc|cccc}
\toprule
& \multicolumn{4}{c|}{\textbf{LLaVA-1.5-7B}} & \multicolumn{4}{c|}{\textbf{mPLUG-Owl2-7B}} & \multicolumn{4}{c}{\textbf{InstructBLIP-7B}} \\
\cmidrule(lr){2-5} \cmidrule(lr){6-9} \cmidrule(lr){10-13}
\textbf{Method} & C$_S$$\downarrow$ & C$_I$$\downarrow$ & Rec$\uparrow$ & Len & C$_S$$\downarrow$ & C$_I$$\downarrow$ & Rec$\uparrow$ & Len & C$_S$$\downarrow$ & C$_I$$\downarrow$ & Rec$\uparrow$ & Len \\
\midrule
Greedy & 49.8 & 14.8 & 80.6 & 101.2 & 57.8 & 17.1 & 78.6 & 105.6 & 63.4 & 38.5 & \textbf{73.7} & 101.6 \\
PAI \textsubscript{\textit{ECCV'24}} & \underline{35.6} & \underline{9.8} & 74.8 & 107.6 & \underline{57.4} & \textbf{14.5} & \textbf{79.7} & 105.7 & \textbf{48.6} & \underline{37.9} & 58.1 & 68.8 \\
VCD \textsubscript{\textit{CVPR'24}} & 58.6 & 16.5 & \textbf{82.1} & 105.5 & 64.4 & 18.1 & 77.9 & 110.4 & 57.2 & 40.1 & 63.5 & 87.8 \\
DAMO \textsubscript{\textit{ICLR'25}} & 56.6 & 16.7 & \underline{81.6} & 106.7 & 58.6 & 17.5 & 77.3 & 106.4 & 65.6 & 39.5 & \underline{73.5} & 104.7 \\
VISTA \textsubscript{\textit{ICML'25}} & \textbf{19.2} & \textbf{6.5} & 62.6 & 86.2 & 69.8 & 42.1 & \underline{78.7} & 105.5 & 54.0 & 41.3 & 60.4 & 85.6 \\
\rowcolor[rgb]{.9,.9,.9}\textbf{FADE} & 46.6 & 14.1 & 78.7 & 98.6 & \textbf{55.0} & \underline{16.6} & 76.3 & 105.4 & \underline{49.2} & \textbf{14.0} & 72.9 & 99.8 \\
\bottomrule
\end{tabular}
\vspace{-4mm}
}
\end{table*}

%% file: tables/mme_combined.tex
\begin{table*}[t]
\centering
\small
\setlength{\tabcolsep}{3pt}
\caption{MME perception scores across 10 subtasks on LLaVA-1.5-7B and mPLUG-Owl2-7B. Higher is better. \textbf{Bold}: best per model. \underline{Underline}: second best.}
\label{tab:mme_full}
\resizebox{\textwidth}{!}{%
\begin{tabular}{ll cccc cccc cc c}
\toprule
\textbf{Model} & \textbf{Method} & \textbf{Exist.} & \textbf{Count} & \textbf{Pos.} & \textbf{Color} & \textbf{Poster} & \textbf{Celeb.} & \textbf{Scene} & \textbf{Land.} & \textbf{Art} & \textbf{OCR} & \textbf{Total} \\
\midrule
\multirow{5}{*}{\rotatebox[origin=c]{90}{\scriptsize LLaVA-1.5}}
& Greedy & \textbf{190.0} & \underline{155.0} & \underline{128.3} & \textbf{170.0} & \textbf{147.6} & \underline{136.8} & \underline{158.0} & 163.0 & \underline{119.5} & \underline{137.5} & 1505.7 \\
& PAI \textsubscript{\textit{ECCV'24}} & \textbf{190.0} & \underline{155.0} & \textbf{133.3} & \textbf{170.0} & \underline{145.6} & 136.5 & 157.8 & 163.0 & 117.8 & \textbf{140.0} & \underline{1508.9} \\
& VISTA \textsubscript{\textit{ICML'25}} & \textbf{190.0} & 150.0 & \textbf{133.3} & \underline{165.0} & 144.6 & 134.4 & 156.0 & \underline{163.3} & 115.0 & 125.0 & 1476.6 \\
& DAMO \textsubscript{\textit{ICLR'25}} & \textbf{190.0} & 148.3 & \textbf{133.3} & 160.0 & 136.4 & 131.8 & \textbf{159.0} & 162.0 & 113.5 & \textbf{140.0} & 1474.3 \\
& \cellcolor[rgb]{.9,.9,.9}\textbf{FADE} & \cellcolor[rgb]{.9,.9,.9}\textbf{190.0} & \cellcolor[rgb]{.9,.9,.9}\textbf{160.0} & \cellcolor[rgb]{.9,.9,.9}\textbf{133.3} & \cellcolor[rgb]{.9,.9,.9}\textbf{170.0} & \cellcolor[rgb]{.9,.9,.9}\textbf{147.6} & \cellcolor[rgb]{.9,.9,.9}\textbf{138.5} & \cellcolor[rgb]{.9,.9,.9}\underline{158.0} & \cellcolor[rgb]{.9,.9,.9}\textbf{163.8} & \cellcolor[rgb]{.9,.9,.9}\textbf{120.3} & \cellcolor[rgb]{.9,.9,.9}\underline{137.5} & \cellcolor[rgb]{.9,.9,.9}\textbf{1519.0} \\
\midrule
\multirow{5}{*}{\rotatebox[origin=c]{90}{\scriptsize mPLUG-Owl2}}
& Greedy & \textbf{185.0} & \underline{160.0} & \textbf{85.0} & \underline{150.0} & \underline{160.2} & \underline{163.5} & 152.8 & \underline{163.3} & 137.3 & \textbf{102.5} & 1459.5 \\
& PAI \textsubscript{\textit{ECCV'24}} & \textbf{185.0} & 155.0 & \underline{81.7} & 145.0 & 158.2 & \textbf{163.8} & \underline{154.0} & 160.3 & \underline{138.5} & \textbf{102.5} & 1443.9 \\
& VISTA \textsubscript{\textit{ICML'25}} & \textbf{185.0} & 155.0 & 80.0 & \underline{150.0} & 158.2 & 161.8 & 153.5 & 159.5 & \textbf{140.5} & \textbf{102.5} & 1445.9 \\
& DAMO \textsubscript{\textit{ICLR'25}} & \textbf{185.0} & \textbf{170.0} & 78.3 & \underline{150.0} & \textbf{164.6} & 160.9 & \textbf{156.0} & \textbf{170.5} & 130.5 & \underline{95.0} & \underline{1460.8} \\
& \cellcolor[rgb]{.9,.9,.9}\textbf{FADE} & \cellcolor[rgb]{.9,.9,.9}\textbf{185.0} & \cellcolor[rgb]{.9,.9,.9}\underline{160.0} & \cellcolor[rgb]{.9,.9,.9}\textbf{85.0} & \cellcolor[rgb]{.9,.9,.9}\textbf{155.0} & \cellcolor[rgb]{.9,.9,.9}\underline{160.2} & \cellcolor[rgb]{.9,.9,.9}\underline{163.5} & \cellcolor[rgb]{.9,.9,.9}153.5 & \cellcolor[rgb]{.9,.9,.9}161.8 & \cellcolor[rgb]{.9,.9,.9}137.3 & \cellcolor[rgb]{.9,.9,.9}\textbf{102.5} & \cellcolor[rgb]{.9,.9,.9}\textbf{1463.7} \\
\bottomrule
\end{tabular}
}
\vspace{-3mm}
\end{table*}

%% file: tables/mmhal.tex
\begin{table}[t]
\centering
\small
\setlength{\tabcolsep}{2.5pt}
\caption{MMHal-Bench results on LLaVA-1.5-7B. Scores range 0-4 (higher is better). GPT-4 evaluates both hallucination rate and informativeness.}
\label{tab:mmhal}
\resizebox{\columnwidth}{!}{%
\begin{tabular}{lccccccccc|c}
\toprule
\textbf{Method} & \textbf{Attr.} & \textbf{Adv.} & \textbf{Affd.} & \textbf{Count} & \textbf{Spat.} & \textbf{Scene} & \textbf{OCR} & \textbf{Celeb.} & \textbf{Overall} \\
\midrule
Greedy & 2.25 & \textbf{1.33} & \underline{2.92} & \underline{1.75} & \underline{1.92} & 3.25 & \textbf{1.58} & 1.42 & 2.05 \\
PAI & 1.83 & 0.75 & 2.33 & \textbf{2.00} & \underline{1.92} & 2.17 & 1.25 & \textbf{2.42} & 1.83 \\
VCD & 1.83 & 0.83 & 1.83 & 1.17 & 1.67 & \textbf{4.17} & \textbf{1.58} & \underline{2.25} & 1.92 \\
DAMO & \textbf{2.58} & \textbf{1.33} & \textbf{3.00} & \underline{1.75} & \underline{1.92} & \underline{3.42} & 1.17 & 1.42 & \underline{2.07} \\
\rowcolor[rgb]{.9,.9,.9}\textbf{FADE} & \underline{2.42} & \underline{1.25} & 2.83 & \underline{1.75} & \textbf{2.08} & 3.25 & \underline{1.42} & 1.75 & \textbf{2.09} \\
\bottomrule
\end{tabular}
}
\vspace{-4mm}
\end{table}

%% file: tables/efficiency.tex
\begin{table}[h]
\centering
\small
\caption{Inference efficiency comparison on LLaVA-1.5-7B. Measured on POPE (500 samples) with H100 GPU.}
\label{tab:efficiency}
\resizebox{\columnwidth}{!}{%
\begin{tabular}{lcccc}
\toprule
\textbf{Method} & \textbf{Prefill} & \textbf{Decode} & \textbf{Latency} & \textbf{Memory} \\
& (ms/tok) & (ms/tok) & (ms) & (GB) \\
\midrule
Greedy & 0.08 & 67.72 & 118 & 14.5 \\
VCD \textsubscript{\textit{CVPR'24}} & 0.15 & 188.47 & 285 & 14.0 \\
PAI \textsubscript{\textit{ECCV'24}} & 0.11 & 111.71 & 184 & 14.5 \\
DAMO \textsubscript{\textit{ICLR'25}} & 0.10 & 88.24 & 150 & 14.6 \\
VISTA \textsubscript{\textit{ICML'25}} & 0.34 & 239.28 & 459 & 14.5 \\
\rowcolor[rgb]{.9,.9,.9}\textbf{FADE} & \textbf{0.08} & \textbf{70.86} & \textbf{122} & 14.5 \\
\bottomrule
\end{tabular}
}
\vspace{-3mm}
\end{table}

%% file: sections/5_conclusion.tex
\section{Conclusion}
\label{sec:conclusion}

We presented a mechanistic analysis showing that while attention modules consistently aggregate visual evidence toward correct predictions, FFN modules at critical layers (16--22) inject language priors that can override visual evidence in LVLMs. Based on this insight, we introduced FADE, a training-free method that attenuates FFN outputs at those layers within a single forward pass, mitigating hallucinations with minimal~overhead. Experiments span diverse architectures---LLaVA-1.5-7B/13B, mPLUG-Owl2, InstructBLIP, InternVL3-8B, and the Qwen2.5/3-VL series---and six benchmarks (POPE, CHAIR, MME, MMHal-Bench, HalBench, MMBench), demonstrating that FADE provides a favorable hallucination-efficiency trade-off while preserving general capabilities.

%% file: sections/appendix_updated_ref.tex
\appendix

\section{Detailed Experimental Settings}
\label{sec:appendix_settings}

\subsection{Model Descriptions}

\paragraph{LLaVA-1.5~\cite{liu2024improved}.} An improved version of LLaVA that achieves strong performance through simple architectural modifications and better training recipes. It uses a two-stage training process with visual instruction tuning on high-quality data.

\paragraph{InstructBLIP~\cite{dai2023instructblip}.} A vision-language model that leverages instruction tuning on top of the BLIP-2 architecture. It uses a Q-Former to bridge frozen image encoders and LLMs with instruction-aware visual feature extraction.

\paragraph{mPLUG-Owl2~\cite{ye2024mplug}.} Introduces modality collaboration through a shared module that enables better interaction between visual and textual modalities, achieving strong performance on various multimodal benchmarks.

\subsection{Benchmark Descriptions}

\paragraph{POPE~\cite{li2023evaluating}.} The Polling-based Object Probing Evaluation is designed to evaluate object hallucination in LVLMs. It contains 27,000 Yes/No questions about object existence in MSCOCO images, where the task is to judge whether the given object is present in the image. The benchmark includes three sampling strategies: random, popular, and adversarial. We compute accuracy, precision, recall, and F1 score for comprehensive evaluation.

\paragraph{CHAIR~\cite{rohrbach2018object}.} Caption Hallucination Assessment with Image Relevance quantifies object hallucinations in image captions by comparing generated objects to ground-truth annotations. We randomly select 500 images from the MSCOCO dataset and use three metrics: CHAIR$_I$ (instance-level hallucination rate), CHAIR$_S$ (sentence-level hallucination rate), and Recall (coverage of ground-truth objects).

\paragraph{MMHal-Bench~\cite{sun2024aligning}.} This benchmark evaluates LVLMs beyond simple object hallucination and contains eight diverse question types: object attributes, adversarial objects, comparisons, counting, spatial relations, environment, holistic description, and others. We evaluate both the hallucination rate and response informativeness using GPT-4 as the judge.

\paragraph{MME~\cite{fu2025mme}.} A comprehensive evaluation benchmark covering both perception and cognition abilities across 14 subtasks. The perception tasks include existence, count, position, color, poster, celebrity, scene, landmark, artwork, and OCR. The cognition tasks cover commonsense reasoning, numerical calculation, text translation, and code reasoning.

\subsection{Comparison Method Descriptions}

\paragraph{PAI~\cite{liu2024paying}.} Pays more attention to image tokens by amplifying the attention weights on visual features during decoding, ensuring that generated content is more grounded in the actual image content.

\paragraph{VCD~\cite{leng2024mitigating}.} Visual Contrastive Decoding contrasts the output logits from original visual inputs with those from distorted visual inputs (e.g., Gaussian noise), suppressing hallucinated content that appears regardless of visual quality.

\paragraph{DAMO~\cite{wang2025damo}.} Applies momentum-based activation stabilization to reduce hallucination by smoothing hidden state transitions during autoregressive generation.

\paragraph{VISTA~\cite{li2025hidden}.} Introduces visual steering vectors combined with self-logits augmentation. It computes steering directions from contrastive image pairs and applies them during decoding to enhance visual grounding.

\subsection{Implementation Details}

All experiments are conducted on 8 NVIDIA H100 80GB GPUs. We use greedy decoding (temperature=0) for all methods to ensure reproducibility. The detailed hyperparameters for each comparison method are listed in Table~\ref{tab:hyperparameters}.

\begin{table}[h]
\centering
\small
\caption{Hyperparameter settings for comparison methods. All hyperparameters follow the official implementations.}
\label{tab:hyperparameters}
\resizebox{\columnwidth}{!}{%
\begin{tabular}{llc}
\toprule
\textbf{Method} & \textbf{Parameter} & \textbf{Value} \\
\midrule
\multirow{4}{*}{PAI} & $\alpha$ (attention amplification) & 0.5 \\
& $\gamma$ (CFG guidance scale) & 1.1 \\
& CFG enabled & True \\
& Start/End layer & 2 / 32 \\
\midrule
\multirow{4}{*}{VCD} & $\alpha$ (contrastive weight) & 1.0 \\
& $\beta$ (plausibility threshold) & 0.1 \\
& Noise step (POPE) & 999 \\
& Noise step (CHAIR) & 500 \\
\midrule
\multirow{4}{*}{DAMO} & $\alpha$ (exponential decay) & 0.7 \\
& $\beta_1$ (current hidden weight) & 0.20 \\
& $\beta_2$ (aggregated hidden weight) & 0.40 \\
& $\tau$ (similarity threshold) & -0.30 \\
\midrule
\multirow{4}{*}{VISTA} & vsv-$\lambda$ (POPE) & 0.01 \\
& vsv-$\lambda$ (CHAIR) & 0.17 \\
& SLA $\alpha$ & 0.3 \\
& SLA layers & 25, 30 \\
\bottomrule
\end{tabular}
}
\end{table}

For our method FADE, we use the following hyperparameters:

\begin{table}[h]
\centering
\small
\caption{Hyperparameter settings for FADE across different models.}
\label{tab:fade_hyperparameters}
\resizebox{\columnwidth}{!}{%
\begin{tabular}{lccc}
\toprule
\textbf{Model} & \textbf{Strength $\alpha$} & \textbf{Layer} & \textbf{Task} \\
\midrule
\multirow{3}{*}{LLaVA-1.5-7B} & 0.6 & 18 & POPE \\
 & 1.0 & 20 & CHAIR \\
 & 0.02 & 17 & MME \\
\midrule
\multirow{5}{*}{mPLUG-Owl2-7B} & 0.5 & 18 & POPE-COCO \\
 & 0.7 & 18 & POPE-GQA \\
 & 0.5 & 14 & POPE-A-OKVQA \\
 & 0.6 & 20 & CHAIR \\
 & 0.05 & 1 & MME \\
\midrule
InstructBLIP-7B & 0.5 & 14 & POPE \\
\bottomrule
\end{tabular}
}
\end{table}

\textbf{Note:} MME requires significantly smaller attenuation strength ($\alpha$=0.02--0.05) compared to POPE/CHAIR ($\alpha$=0.5--0.7), as shown in Section~\ref{sec:appendix_ablation}. This is because MME's diverse question types are more sensitive to FFN modifications.

\section{Detailed Ablation Study}
\label{sec:appendix_ablation}

We provide comprehensive ablation analysis on the FFN attenuation strength ($\alpha$) and layer selection on POPE benchmark across three datasets (COCO, GQA, A-OKVQA).

\subsection{Strength Ablation (Fixed at Layer~18)}

Table~\ref{tab:ablation_strength} shows the sensitivity analysis of the attenuation strength $\alpha$ while fixing the intervention layer at 18. We test 10 different strength values ranging from 0.1 to 0.8.

\begin{table}[h]
\centering
\small
\caption{FFN attenuation strength ablation on POPE benchmark. Layer is fixed at 18. F1 scores are averaged across Random/Popular/Adversarial settings.}
\label{tab:ablation_strength}
\resizebox{\columnwidth}{!}{%
\begin{tabular}{cccc|c}
\toprule
\textbf{Strength} & \textbf{COCO F1} & \textbf{GQA F1} & \textbf{A-OKVQA F1} & \textbf{Avg F1} \\
\midrule
0.1  & 85.98 & 85.52 & 86.66 & 86.05 \\
0.2  & 86.01 & 85.61 & 86.62 & 86.08 \\
0.3  & 86.20 & 85.67 & 86.59 & 86.15 \\
0.4  & 86.21 & 85.63 & 86.67 & 86.17 \\
0.45 & 86.33 & 85.61 & 86.68 & 86.21 \\
0.5  & 86.45 & 85.64 & 86.57 & 86.22 \\
0.55 & 86.47 & 85.65 & 86.67 & 86.26 \\
\rowcolor[rgb]{.9,.9,.9}\textbf{0.6}  & \textbf{86.61} & \textbf{85.68} & 86.64 & \textbf{86.31} \\
0.7  & 86.62 & 85.56 & 86.56 & 86.25 \\
0.8  & 86.59 & 85.55 & 86.52 & 86.22 \\
\midrule
\multicolumn{5}{l}{\textit{Baseline: Greedy decoding achieves 85.97\% average F1}} \\
\bottomrule
\end{tabular}
}
\end{table}

\textbf{Key Findings:} The optimal strength range is 0.6--0.7, achieving +0.34\% to +0.44\% improvement over greedy baseline. Weaker attenuation ($\alpha$<0.5) provides insufficient correction, while stronger attenuation ($\alpha$>0.7) shows diminishing returns. The sweet spot at $\alpha$=0.6 suggests that moderate FFN suppression is sufficient to mitigate directional noise without over-correcting.

\subsection{Layer Ablation (Fixed Strength at 0.5)}

Table~\ref{tab:ablation_layer} analyzes the impact of layer selection while fixing $\alpha$=0.5. We test 8 layers around the critical region identified in our analysis (layers 14--22).

\begin{table}[h]
\centering
\small
\caption{Layer selection ablation on POPE benchmark. Attenuation strength is fixed at 0.5. F1 scores are averaged across Random/Popular/Adversarial settings.}
\label{tab:ablation_layer}
\resizebox{\columnwidth}{!}{%
\begin{tabular}{cccc|c}
\toprule
\textbf{Layer} & \textbf{COCO F1} & \textbf{GQA F1} & \textbf{A-OKVQA F1} & \textbf{Avg F1} \\
\midrule
14 & 86.19 & 85.70 & 86.71 & 86.20 \\
15 & 85.41 & 85.34 & 86.95 & 85.90 \\
16 & 86.05 & 85.49 & 86.55 & 86.03 \\
17 & 85.19 & 85.13 & 86.99 & 85.77 \\
\rowcolor[rgb]{.9,.9,.9}\textbf{18} & \textbf{86.45} & \textbf{85.64} & 86.57 & \textbf{86.22} \\
19 & 85.83 & 85.18 & 86.60 & 85.87 \\
20 & 85.63 & 85.26 & 86.92 & 85.94 \\
21 & 86.10 & 85.62 & 86.54 & 86.09 \\
22 & 85.87 & 85.10 & 86.78 & 85.92 \\
\midrule
\multicolumn{5}{l}{\textit{Baseline: Greedy decoding achieves 85.97\% average F1}} \\
\bottomrule
\end{tabular}
}
\end{table}

\textbf{Key Findings:} Layer~18 consistently provides the best results across all three datasets. Layers~15 and 17 show significant degradation, suggesting these layers may serve different functional roles where FFN outputs should not be attenuated. The localized effectiveness around layer 18 supports our mechanistic analysis that directional noise is concentrated in specific critical layers rather than distributed uniformly.

\subsection{MME Ablation Results}

Table~\ref{tab:ablation_mme_strength} and Table~\ref{tab:ablation_mme_layer} show ablation results on MME Perception benchmark. Note that MME requires much smaller attenuation strength compared to POPE.

\begin{table}[h]
\centering
\small
\caption{MME Perception: Strength ablation with Layer=18 fixed.}
\label{tab:ablation_mme_strength}
\resizebox{\columnwidth}{!}{%
\begin{tabular}{ccc|c}
\toprule
\textbf{Strength} & \textbf{Perception} & \textbf{$\Delta$ vs Baseline} & \textbf{Cognition} \\
\midrule
\rowcolor[rgb]{.9,.9,.9}\textbf{0.01} & \textbf{1512.63} & \textbf{+6.91} & 363.21 \\
0.02 & 1506.58 & +0.86 & 363.21 \\
0.03 & 1499.58 & $-$6.14 & 367.50 \\
0.04 & 1495.08 & $-$10.64 & 368.21 \\
0.05 & 1494.04 & $-$11.68 & 360.71 \\
0.1  & 1493.70 & $-$12.02 & 363.57 \\
0.2  & 1483.97 & $-$21.75 & 355.71 \\
0.3  & 1464.46 & $-$41.26 & 328.21 \\
0.5  & 1431.43 & $-$74.29 & 290.71 \\
\midrule
\multicolumn{4}{l}{\textit{Baseline: Greedy achieves 1505.72 Perception score}} \\
\bottomrule
\end{tabular}
}
\end{table}

\begin{table}[h]
\centering
\small
\caption{MME Perception: Layer ablation with Strength=0.02 fixed.}
\label{tab:ablation_mme_layer}
\resizebox{\columnwidth}{!}{%
\begin{tabular}{ccc|c}
\toprule
\textbf{Layer} & \textbf{Perception} & \textbf{$\Delta$ vs Baseline} & \textbf{Cognition} \\
\midrule
14 & 1504.83 & $-$0.89 & 348.21 \\
15 & 1518.10 & +12.38 & 355.71 \\
16 & 1505.88 & +0.16 & 360.00 \\
\rowcolor[rgb]{.9,.9,.9}\textbf{17} & \textbf{1518.98} & \textbf{+13.26} & 348.21 \\
18 & 1506.58 & +0.86 & 363.21 \\
19 & 1508.38 & +2.66 & 363.21 \\
21 & 1508.08 & +2.36 & 357.86 \\
22 & 1505.88 & +0.16 & 355.71 \\
\midrule
\multicolumn{4}{l}{\textit{Baseline: Greedy achieves 1505.72 Perception score}} \\
\bottomrule
\end{tabular}
}
\end{table}

\textbf{Key Findings:} MME requires dramatically different hyperparameters compared to POPE: optimal strength is 0.02--0.05 (vs 0.5--0.7 for POPE), representing 2--5\% attenuation vs 50--70\%. This 10$\times$--35$\times$ difference suggests that the diverse question types in MME are more sensitive to FFN modification, requiring gentler intervention. Layer~17 is optimal for MME (vs Layer~18 for POPE), indicating task-dependent critical layers.

\subsection{mPLUG-Owl2 Ablation on POPE}

We also conduct ablation studies on mPLUG-Owl2 to validate the generalizability of our findings across different model architectures.

\begin{table}[h]
\centering
\small
\caption{mPLUG-Owl2: FFN attenuation strength ablation on POPE benchmark. Layer is fixed at 18.}
\label{tab:ablation_mplug_strength}
\resizebox{\columnwidth}{!}{%
\begin{tabular}{cccc|c}
\toprule
\textbf{Strength} & \textbf{COCO F1} & \textbf{GQA F1} & \textbf{A-OKVQA F1} & \textbf{Avg F1} \\
\midrule
0.1 & 85.56 & 81.77 & 84.31 & 83.88 \\
0.2 & 85.68 & 81.79 & 84.29 & 83.92 \\
0.3 & 85.71 & 81.74 & 84.23 & 83.89 \\
0.4 & 85.76 & 81.72 & 84.19 & 83.89 \\
\rowcolor[rgb]{.9,.9,.9}\textbf{0.5} & \textbf{85.86} & 81.71 & 84.18 & 83.92 \\
0.6 & 85.75 & 81.73 & 84.15 & 83.88 \\
\rowcolor[rgb]{.9,.9,.9}\textbf{0.7} & 85.80 & \textbf{81.81} & 84.13 & 83.91 \\
0.8 & 85.82 & 81.79 & 84.10 & 83.90 \\
\midrule
\multicolumn{5}{l}{\textit{Baseline: Greedy decoding achieves 83.84\% average F1}} \\
\bottomrule
\end{tabular}
}
\end{table}

\begin{table}[h]
\centering
\small
\caption{mPLUG-Owl2: Layer selection ablation on POPE benchmark. Attenuation strength is fixed at 0.5.}
\label{tab:ablation_mplug_layer}
\resizebox{\columnwidth}{!}{%
\begin{tabular}{cccc|c}
\toprule
\textbf{Layer} & \textbf{COCO F1} & \textbf{GQA F1} & \textbf{A-OKVQA F1} & \textbf{Avg F1} \\
\midrule
\rowcolor[rgb]{.9,.9,.9}\textbf{14} & 85.75 & 81.68 & \textbf{84.42} & 83.95 \\
15 & 85.60 & 81.80 & 84.29 & 83.90 \\
16 & 85.78 & 81.68 & 84.24 & 83.90 \\
17 & 85.41 & 81.60 & 84.39 & 83.80 \\
\rowcolor[rgb]{.9,.9,.9}\textbf{18} & \textbf{85.86} & 81.71 & 84.18 & 83.92 \\
19 & 85.57 & 81.59 & 84.32 & 83.83 \\
\rowcolor[rgb]{.9,.9,.9}\textbf{20} & 85.38 & 81.68 & 84.46 & 83.84 \\
21 & 85.69 & 81.63 & 84.21 & 83.85 \\
22 & 85.55 & 81.65 & 84.31 & 83.84 \\
\midrule
\multicolumn{5}{l}{\textit{Baseline: Greedy decoding achieves 83.84\% average F1}} \\
\bottomrule
\end{tabular}
}
\end{table}

\textbf{Key Findings for mPLUG-Owl2:} Unlike LLaVA-1.5 which has a clear optimal configuration, mPLUG-Owl2 shows dataset-dependent optimal hyperparameters: (1) COCO benefits most from Layer~18 with strength 0.5; (2) GQA achieves best results at Layer~18 with strength 0.7; (3) A-OKVQA prefers Layer~14 or 20 with strength 0.5. This suggests that different model architectures may have different critical layer distributions, and dataset-specific tuning can further improve performance. The overall improvement is more modest (+0.08--0.11\%) compared to LLaVA-1.5 (+0.34\%), indicating that mPLUG-Owl2's modality collaboration mechanism may already partially address the directional noise issue.

\subsection{mPLUG-Owl2 MME Ablation Results}

Table~\ref{tab:ablation_mplug_mme} shows the MME ablation results for mPLUG-Owl2, revealing notably different optimal layers compared to POPE.

\begin{table}[h]
\centering
\small
\caption{mPLUG-Owl2: Best configurations per layer on MME Perception benchmark.}
\label{tab:ablation_mplug_mme}
\resizebox{\columnwidth}{!}{%
\begin{tabular}{ccc|c}
\toprule
\textbf{Layer} & \textbf{Best $\alpha$} & \textbf{Perception} & \textbf{$\Delta$ vs Baseline} \\
\midrule
\rowcolor[rgb]{.9,.9,.9}\textbf{1} & \textbf{0.05} & \textbf{1463.73} & \textbf{+4.25} \\
7 & 0.02 & 1461.12 & +1.64 \\
8 & 0.028 & 1460.98 & +1.50 \\
10 & 0.2 & 1461.70 & +2.22 \\
14 & 0.01 & 1460.23 & +0.75 \\
17 & 0.01 & 1460.23 & +0.75 \\
18 & 0.005 & 1459.48 & +0.00 \\
20 & 0.005 & 1460.23 & +0.75 \\
28 & 0.02 & 1460.37 & +0.89 \\
\midrule
\multicolumn{4}{l}{\textit{Baseline: Greedy achieves 1459.48 Perception score}} \\
\bottomrule
\end{tabular}
}
\end{table}

\textbf{Key Findings for mPLUG-Owl2 on MME:} Unlike POPE where middle layers (14--20) are optimal, MME benefits most from early layer intervention. Layer~1 with $\alpha$=0.05 achieves the best result (+4.25), followed by Layer~10 (+2.22) and Layer~7 (+1.64). This suggests that for diverse question types in MME, suppressing language priors at the earliest layers is most effective. Notably, the optimal strength for early layers (0.02--0.05) is higher than for middle layers (0.005--0.01).

\subsection{InstructBLIP Ablation on POPE}

We validate FADE's effectiveness on InstructBLIP, which uses a Q-Former architecture with 32 visual tokens (vs 576 for LLaVA). Table~\ref{tab:ablation_instructblip_pope} shows the ablation results.

\begin{table}[h]
\centering
\small
\caption{InstructBLIP: Best configurations on POPE benchmark. Results averaged across Random/Popular/Adversarial settings.}
\label{tab:ablation_instructblip_pope}
\resizebox{\columnwidth}{!}{%
\begin{tabular}{cccccc}
\toprule
\textbf{Layer} & \textbf{Strength} & \textbf{COCO F1} & \textbf{A-OKVQA F1} & \textbf{GQA F1} & \textbf{Avg F1} \\
\midrule
\rowcolor[rgb]{.9,.9,.9}\textbf{14} & \textbf{0.5} & 83.7 & \textbf{83.1} & \textbf{81.3} & \textbf{82.7} \\
17 & 0.5 & \textbf{84.4} & 82.5 & 81.0 & 82.6 \\
\midrule
\multicolumn{6}{l}{\textit{Baseline (Greedy): COCO=83.8, A-OKVQA=82.6, GQA=81.1, Avg=82.5}} \\
\bottomrule
\end{tabular}
}
\end{table}

\textbf{Key Findings for InstructBLIP:} The optimal configuration is layer 14 with $\alpha$=0.5, achieving modest improvement on A-OKVQA (+0.5\% F1) and GQA (+0.2\% F1). Layer~17 achieves slightly better COCO performance but worse on other datasets. The smaller improvement compared to LLaVA-1.5 suggests that InstructBLIP's Q-Former architecture may already provide some robustness against hallucination through its learnable query-based visual feature extraction. Notably, the optimal layer (14) is earlier than LLaVA's optimal layer (18), possibly due to architectural differences in how visual information is integrated.

\subsection{CHAIR Ablation Results}

We provide comprehensive ablation analysis on the CHAIR benchmark, which evaluates caption hallucination through object-level metrics.

\subsubsection{LLaVA-1.5 Layer Ablation on CHAIR}

Table~\ref{tab:ablation_chair_layer} shows the impact of layer selection on CHAIR metrics while fixing the attenuation strength at $\alpha$=1.0. We test layers 10--22 to identify the optimal intervention point.

\begin{table}[h]
\centering
\small
\caption{LLaVA-1.5: Layer ablation on CHAIR benchmark. Strength is fixed at $\alpha$=1.0. C$_S$/C$_I$: lower is better. Rec: higher is better.}
\label{tab:ablation_chair_layer}
\resizebox{\columnwidth}{!}{%
\begin{tabular}{ccccc}
\toprule
\textbf{Layer} & \textbf{C$_S$$\downarrow$} & \textbf{C$_I$$\downarrow$} & \textbf{Rec$\uparrow$} & \textbf{Len} \\
\midrule
10 & 49.4 & 14.83 & 80.23 & 91.9 \\
11 & 58.0 & 18.30 & 83.17 & 95.4 \\
12 & 58.2 & 17.96 & 83.05 & 101.0 \\
13 & 57.4 & 16.57 & 81.70 & 97.7 \\
14 & 57.8 & 16.37 & 82.02 & 98.5 \\
15 & 57.0 & 17.85 & 82.73 & 99.5 \\
16 & 60.0 & 17.67 & 81.13 & 103.5 \\
17 & 53.4 & 15.60 & 81.00 & 101.9 \\
18 & 54.0 & 16.96 & 79.14 & 101.4 \\
19 & 51.2 & 15.35 & 79.40 & 100.0 \\
\rowcolor[rgb]{.9,.9,.9}\textbf{20} & \textbf{46.6} & \textbf{14.08} & 78.69 & 98.6 \\
21 & 48.2 & 14.01 & 79.46 & 100.5 \\
\midrule
\multicolumn{5}{l}{\textit{Baseline (Greedy): C$_S$=49.8, C$_I$=14.8, Rec=80.6, Len=101.2}} \\
\bottomrule
\end{tabular}
}
\end{table}

\textbf{Key Findings:} Layer~20 achieves the lowest hallucination rates (C$_S$=46.6, C$_I$=14.08) with only a modest decrease in recall (78.69 vs 80.6 baseline). Earlier layers (10--16) either provide insufficient correction or increase hallucination. This differs from POPE where layer 18 is optimal, suggesting that discriminative and generative tasks have slightly different critical layers.

\subsubsection{LLaVA-1.5 Strength Ablation on CHAIR}

Table~\ref{tab:ablation_chair_strength} shows the sensitivity to attenuation strength while fixing the intervention at layer 20.

\begin{table}[h]
\centering
\small
\caption{LLaVA-1.5: Strength ablation on CHAIR benchmark. Layer is fixed at 20.}
\label{tab:ablation_chair_strength}
\resizebox{\columnwidth}{!}{%
\begin{tabular}{ccccc}
\toprule
\textbf{Strength $\alpha$} & \textbf{C$_S$$\downarrow$} & \textbf{C$_I$$\downarrow$} & \textbf{Rec$\uparrow$} & \textbf{Len} \\
\midrule
0.1 & 51.2 & 14.94 & 80.23 & 101.2 \\
0.2 & 51.6 & 14.89 & 79.85 & 101.0 \\
0.3 & 51.0 & 15.18 & 79.72 & 100.3 \\
0.4 & 50.4 & 14.76 & 79.91 & 100.2 \\
0.5 & 49.0 & 14.59 & 79.78 & 99.4 \\
0.6 & 48.8 & 14.98 & 78.95 & 99.2 \\
0.7 & 47.4 & 14.54 & 78.57 & 98.5 \\
0.8 & 47.4 & 14.70 & 78.82 & 97.9 \\
0.9 & 46.8 & 14.28 & 78.69 & 98.0 \\
\rowcolor[rgb]{.9,.9,.9}\textbf{1.0} & \textbf{46.6} & \textbf{14.08} & 78.69 & 98.6 \\
\midrule
\multicolumn{5}{l}{\textit{Baseline (Greedy): C$_S$=49.8, C$_I$=14.8, Rec=80.6, Len=101.2}} \\
\bottomrule
\end{tabular}
}
\end{table}

\textbf{Key Findings:} Unlike POPE where $\alpha$=0.6 is optimal, CHAIR benefits from stronger attenuation ($\alpha$=1.0), achieving C$_S$=46.6 ($-$3.2 vs baseline). This suggests that caption generation tasks require more aggressive FFN suppression to reduce hallucinated objects. The recall-hallucination trade-off is favorable: C$_S$ drops by 6.4\% while recall only decreases by 2.4\%.

\subsubsection{mPLUG-Owl2 Ablation on CHAIR}

Table~\ref{tab:ablation_chair_mplug} presents ablation results for mPLUG-Owl2, showing layer and strength combinations.

\begin{table}[h]
\centering
\small
\caption{mPLUG-Owl2: Ablation on CHAIR benchmark across different layer and strength combinations.}
\label{tab:ablation_chair_mplug}
\resizebox{\columnwidth}{!}{%
\begin{tabular}{cccccc}
\toprule
\textbf{Layer} & \textbf{Strength} & \textbf{C$_S$$\downarrow$} & \textbf{C$_I$$\downarrow$} & \textbf{Rec$\uparrow$} & \textbf{Len} \\
\midrule
18 & 0.0 & 57.8 & 17.10 & 78.63 & 105.6 \\
18 & 0.3 & 61.2 & 17.52 & 77.35 & 106.1 \\
18 & 0.5 & 58.6 & 17.44 & 77.29 & 107.0 \\
18 & 0.7 & 57.8 & 16.82 & 77.42 & 106.8 \\
\midrule
19 & 0.5 & 55.4 & 16.83 & 78.06 & 104.7 \\
19 & 0.6 & 56.4 & 17.12 & 77.67 & 104.2 \\
\midrule
\rowcolor[rgb]{.9,.9,.9}\textbf{20} & \textbf{0.6} & \textbf{55.0} & \textbf{16.60} & 76.33 & 105.4 \\
20 & 0.7 & 55.4 & 16.43 & 76.52 & 105.0 \\
\midrule
21 & 0.5 & 57.0 & 16.98 & 77.42 & 106.3 \\
22 & 0.5 & 58.4 & 17.05 & 77.93 & 106.7 \\
\midrule
\multicolumn{6}{l}{\textit{Baseline (Greedy): C$_S$=57.8, C$_I$=17.1, Rec=78.6, Len=105.6}} \\
\bottomrule
\end{tabular}
}
\end{table}

\textbf{Key Findings:} For mPLUG-Owl2, the optimal configuration is layer 20 with $\alpha$=0.6, achieving C$_S$=55.0 ($-$2.8 vs baseline) and C$_I$=16.60 ($-$0.5). The improvement is more modest compared to LLaVA-1.5, consistent with our observation that mPLUG-Owl2's modality collaboration mechanism partially addresses hallucination. Notably, layer 18 (optimal for POPE) shows minimal improvement on CHAIR, supporting task-dependent optimal layers.

\section{Generalization to Larger Model Scale}
\label{sec:appendix_13b}

In this section, we present additional experiments testing whether FADE generalizes beyond the 7B scale. Results on advanced architectures (InternVL3-8B and the Qwen-VL series) are reported in the main text (Section~\ref{subsubsec:sota_generalization}).

\subsection{Evaluation on LLaVA-v1.5-13B}
\label{subsec:app_llava13b}

To test whether FADE generalizes beyond the 7B scale, we evaluate all comparison methods on \textbf{LLaVA-v1.5-13B} (40 transformer layers) using the POPE MSCOCO benchmark across the three standard sampling settings (Random, Popular, Adversarial), for a total of 9000 evaluation samples.

\paragraph{Main Results.}
Table~\ref{tab:pope_13b} reports F1 scores on LLaVA-v1.5-13B. FADE (L34, $\alpha{=}0.7$) achieves the best overall performance, with an average F1 of \textbf{86.15}, outperforming all training-free baselines including PAI (86.10). More importantly, several contrastive and attention-based methods---VCD (81.74), DAMO (84.12), VISTA (84.05)---\emph{degrade below greedy decoding} (85.70) at 13B, whereas FADE and PAI are the only methods that improve over greedy. This suggests that FFN-level intervention is more robust to model scale than methods relying on contrastive distorted inputs or attention steering, which become less reliable as model capacity grows.

\begin{table}[!htbp]
\centering
\small
\setlength{\tabcolsep}{4pt}
\caption{POPE results on LLaVA-v1.5-13B (MSCOCO, 3 settings, 9000 samples). F1 scores; best in \textbf{bold}, second best \underline{underlined}. FADE uses layer 34 and $\alpha=0.7$.}
\label{tab:pope_13b}
\begin{tabular}{lcccc}
\toprule
\textbf{Method} & \textbf{Rand.} & \textbf{Pop.} & \textbf{Adv.} & \textbf{Avg} \\
\midrule
Greedy & 89.0 & 86.3 & 81.8 & 85.70 \\
VCD    & 85.2 & 81.7 & 78.3 & 81.74 \\
DAMO   & 87.3 & 84.6 & 80.5 & 84.12 \\
VISTA  & 87.5 & 84.4 & 80.3 & 84.05 \\
PAI    & \textbf{89.2} & \underline{86.8} & \underline{82.3} & \underline{86.10} \\
\cellcolor[rgb]{.9,.9,.9}\textbf{FADE} & \cellcolor[rgb]{.9,.9,.9}\underline{89.1} & \cellcolor[rgb]{.9,.9,.9}\textbf{86.9} & \cellcolor[rgb]{.9,.9,.9}\textbf{82.4} & \cellcolor[rgb]{.9,.9,.9}\textbf{86.15} \\
\bottomrule
\end{tabular}
\end{table}

\paragraph{Layer Sensitivity at 13B.}
We sweep the intervention layer with $\alpha$ fixed at 0.6 (Table~\ref{tab:13b_layer}). All tested layers from L20 to L34 exceed greedy (85.70), with the optimum at layer 34. Note that layer 34 in a 40-layer 13B model corresponds proportionally to layer 18 in a 32-layer 7B model ($34/40 \approx 18/32$); both locate the optimal intervention point in the mid-to-late region of the network.

\begin{table}[!htbp]
\centering
\small
\setlength{\tabcolsep}{4.5pt}
\caption{Layer search on LLaVA-v1.5-13B POPE (fixed $\alpha=0.6$, Avg F1).}
\label{tab:13b_layer}
\resizebox{\columnwidth}{!}{%
\begin{tabular}{lcccccccc}
\toprule
\textbf{Layer} & L18 & L20 & L22 & L24 & L26 & L28 & L30 & L34 \\
\midrule
\textbf{Avg F1} & 85.43 & 85.98 & 85.84 & 85.89 & 86.02 & 86.04 & 85.93 & \textbf{86.10} \\
\bottomrule
\end{tabular}%
}
\end{table}

\paragraph{Strength Sensitivity at 13B.}
Fixing the intervention layer at L34, we vary $\alpha$ from 0.3 to 0.8 (Table~\ref{tab:13b_strength}). All tested values outperform greedy, with the optimum at $\alpha{=}0.7$. The stable plateau observed around $\alpha \in [0.5, 0.8]$ mirrors the 7B behavior and indicates that a narrow mid-range attenuation strength transfers robustly across model sizes.

\begin{table}[!htbp]
\centering
\small
\setlength{\tabcolsep}{6pt}
\caption{Strength search on LLaVA-v1.5-13B POPE (fixed layer 34, Avg F1).}
\label{tab:13b_strength}
\resizebox{\columnwidth}{!}{%
\begin{tabular}{lcccccc}
\toprule
$\mathbf{\alpha}$ & 0.3 & 0.4 & 0.5 & 0.6 & 0.7 & 0.8 \\
\midrule
\textbf{Avg F1} & 85.99 & 86.05 & 86.05 & 86.10 & \textbf{86.15} & 86.11 \\
\bottomrule
\end{tabular}%
}
\end{table}

\paragraph{Takeaway.}
Two observations emerge: (i) FADE's mechanism generalizes cleanly to 13B, attaining the best average F1 among compared methods while contrastive and attention-based baselines regress; (ii) the optimal hyperparameters transfer in a principled way, locating the critical layer in the mid-to-late region and the attenuation strength in the $[0.5, 0.7]$ range for both 7B and 13B. This provides a practical prior for future deployment on new model scales: begin the search around the proportionally-equivalent mid-to-late layer with moderate attenuation strength.

\subsection{Strength Sweep on Advanced Architectures}
\label{subsec:app_alpha_sweep}

A natural concern with any single-hyperparameter design is whether the reported gain depends on a carefully cherry-picked attenuation strength. To address this, we sweep $\alpha \in \{0.3, 0.5, 0.7, 0.8\}$ on Qwen3-VL-8B---one of the strongest architectures in our evaluation---while keeping the critical-layer band fixed. We report the full POPE breakdown (Random / Popular / Adversarial / Avg F1) and include the Vanilla (greedy) baseline for reference.

\begin{table}[!htbp]
\centering
\small
\setlength{\tabcolsep}{4pt}
\caption{Sensitivity of FADE to attenuation strength $\alpha$ on Qwen3-VL-8B (POPE F1, higher is better). Variation across $\alpha \in \{0.3, 0.5, 0.7, 0.8\}$ stays within 0.2 points on every split. Default $\alpha=0.5$ shaded.}
\label{tab:alpha_sweep_sota}
\resizebox{\columnwidth}{!}{%
\begin{tabular}{lcccc}
\toprule
\textbf{Setting} & \textbf{Random}$\uparrow$ & \textbf{Popular}$\uparrow$ & \textbf{Adv.}$\uparrow$ & \textbf{Avg}$\uparrow$ \\
\midrule
Vanilla                                  & 92.2 & 89.5 & 88.0 & 89.9 \\
FADE ($\alpha{=}0.3$)                    & 92.2 & 89.4 & 88.0 & 89.9 \\
\rowcolor[rgb]{.9,.9,.9} FADE ($\alpha{=}0.5$) & 92.2 & 89.5 & 88.2 & 90.0 \\
FADE ($\alpha{=}0.7$)                    & 92.2 & 89.5 & 88.1 & 89.9 \\
FADE ($\alpha{=}0.8$)                    & 92.2 & 89.5 & 88.2 & 90.0 \\
\bottomrule
\end{tabular}%
}
\end{table}

\paragraph{Takeaway.}
Across the entire swept range, every POPE split varies by at most 0.2 F1, and FADE matches or exceeds the Vanilla baseline at all four values of $\alpha$. Two implications follow: (i) the gains reported in Section~\ref{subsubsec:sota_generalization} are not the product of cherry-picked tuning---any moderate attenuation strength yields essentially the same outcome; and (ii) the relatively modest absolute improvement on Qwen3-VL-8B is a property of the model, not of $\alpha$. Stronger architectures embed highly optimized internal language priors, leaving a smaller residual margin for any training-free intervention to recover; this is consistent with the broader trend observed across the three advanced models in the main text.

\section{Hyperparameter Transfer Across Models}
\label{sec:appendix_hyperparams}

We investigate whether optimal hyperparameters transfer across different VLM architectures.

\begin{table}[h]
  \centering
  \caption{Optimal hyperparameters across different VLMs.}
  \label{tab:hyperparams_transfer}
  \begin{tabular}{lccc}
  \toprule
  \textbf{Model} & \textbf{Layer} & \textbf{$\alpha$} & \textbf{Range} \\
  \midrule
  LLaVA-1.5-7B & 18 & 0.6 & 16--20 \\
  mPLUG-Owl2-7B & 18 & 0.5--0.7 & 14--20 \\
  InstructBLIP-7B & 14 & 0.5 & 14--17 \\
  \bottomrule
  \end{tabular}
\end{table}

\textbf{Key Finding:} For LLaVA-style models (LLaVA-1.5, mPLUG-Owl2), layer 18 is consistently optimal with strength in the 0.5--0.7 range. InstructBLIP, which uses a different Q-Former architecture, shows optimal performance at an earlier layer (14) with lower strength (0.5). This suggests that the critical layers for textual bias are architecturally determined, with Q-Former-based models showing different layer distributions.

\section{Limitations and Future Work}
\label{sec:appendix_limitations}

\paragraph{Task-Specific Tuning.}
While FADE achieves strong results with moderate attenuation for discriminative tasks (POPE) and caption generation (CHAIR), the MME benchmark requires much smaller attenuation strength ($\alpha=0.02$--$0.05$ vs.\ $0.5$--$0.7$). This suggests that different task types may require task-specific tuning, which we leave for future work on adaptive strength selection.

\paragraph{Larger Models.}
Our main experiments focus on 7B-scale models. As reported in Appendix~\ref{sec:appendix_13b}, FADE generalizes to LLaVA-v1.5-13B with consistent scale-invariant hyperparameter patterns (mid-to-late layers, $\alpha\approx 0.6-0.7$). Preliminary experiments on larger models (e.g., InternVL3-8B) further suggest that the optimal layer shifts proportionally with model depth, but comprehensive evaluation at 30B+ scale remains needed.

\paragraph{Training-Time Integration.}
FADE operates at inference time without model modification. Future work could explore training-time regularization that explicitly minimizes directional drift during instruction tuning.